\documentclass[final]{cvpr}
\usepackage[utf8]{inputenc} % allow utf-8 input
\usepackage[T1]{fontenc}    % use 8-bit T1 fonts
\usepackage{microtype}      % microtypography

\usepackage{subcaption}
\usepackage{wrapfig}
\usepackage{float}
\usepackage{chngpage}
\usepackage{enumitem}

\usepackage{graphicx}
\usepackage{amsmath}
\usepackage[
  pagebackref,
  breaklinks=true,
  letterpaper=true,
  pageanchor=true,
  plainpages=false,
  pdfpagelabels,
  bookmarks=false,
  bookmarksnumbered,
  colorlinks=true,
]{hyperref}
\usepackage{mathtools}
\usepackage{microtype}
\usepackage{url}
\usepackage{amsfonts}
\usepackage{amssymb}
\usepackage{amsthm}
\usepackage{mathrsfs}
\usepackage{verbatim}
\usepackage{booktabs}
\usepackage{multirow}
\usepackage{footnote}
\usepackage{blindtext}
\usepackage{mwe}
\usepackage{adjustbox}
\usepackage{algorithm}
\usepackage[linesnumbered,ruled,algo2e]{algorithm2e}
\usepackage[font=small,labelfont=bf]{caption}

\makesavenoteenv{tabular}
\usepackage[toc,page]{appendix}
\usepackage{comment}

\renewcommand{\ie}{\textit{i.e.,}}

\newcommand{\ABS}[1]{\left| #1 \right|}

\DeclareMathOperator*{\argmin}{argmin}

\usepackage{times}
\usepackage{epsfig}
\usepackage{graphicx}
\usepackage{amsmath}
\usepackage{amssymb}

\def\cvprPaperID{2061} % *** Enter the CVPR Paper ID here
\def\confYear{CVPR 2021}

\begin{document}

\title{Meta Pseudo Labels}

\author{Hieu Pham, Zihang Dai, Qizhe Xie, Minh-Thang Luong, Quoc V. Le\\
Google AI, Brain Team, Mountain View, CA 94043\\
{\tt\small\{hyhieu,zihangd,qizhex,thangluong,qvl\}@google.com}
}

\maketitle
\begin{abstract}
We present Meta Pseudo Labels, a semi-supervised learning method that achieves a new state-of-the-art top-1 accuracy of 90.2\% on ImageNet, which is 1.6\% better than the existing state-of-the-art~\cite{foret2020sharpnessaware}. Like Pseudo Labels, Meta Pseudo Labels has a teacher network to generate pseudo labels on unlabeled data to teach a student network. However, unlike Pseudo Labels where the teacher is fixed,  the teacher in Meta Pseudo Labels is constantly adapted by the feedback of the student's performance on the labeled dataset. As a result, the teacher generates better pseudo labels to teach the student.\footnote{Code is available at \url{https://github.com/google-research/google-research/tree/master/meta_pseudo_labels}.}

% Meta Pseudo Labels is based on Pseudo Labels in that it has a teacher network to generate pseudo labels on unlabeled data to teach a student network. However, in Meta Pseudo Labels, the teacher is not fixed but constantly adapted by the feedback of how well the student performs on the labeled dataset. As a result, the teacher generates better pseudo labels to teach the student.

\end{abstract}

\section{\label{sec:intro}Introduction}
\begin{figure*}[htb!]
  \centering
  \includegraphics[width=1\linewidth]{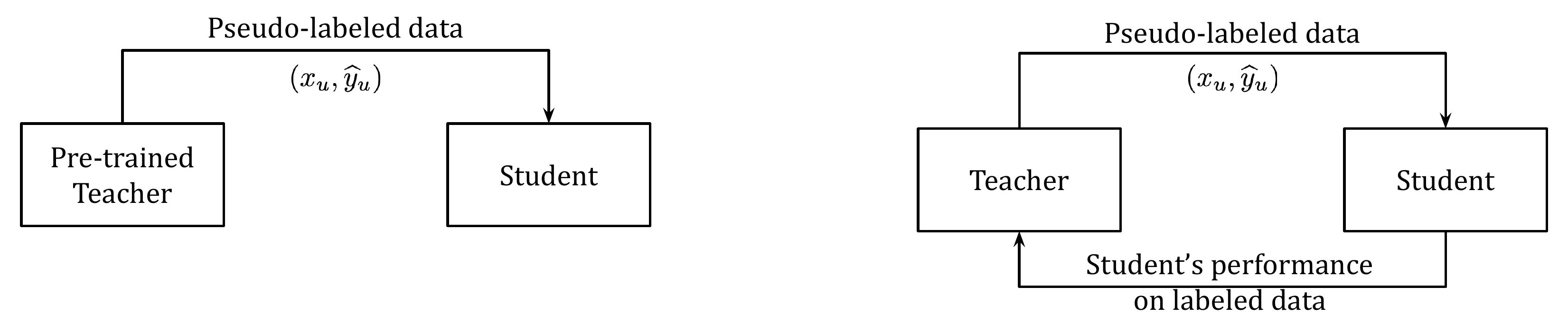}
  \caption{\label{fig:mpl_process}The difference between Pseudo Labels and Meta Pseudo Labels. \textbf{Left:} Pseudo Labels, where a fixed pre-trained teacher generates pseudo labels for the student to learn from. \textbf{Right:} Meta Pseudo Labels, where the teacher is trained along with the student. The student is trained based on the pseudo labels generated by the teacher (top arrow). The teacher is trained based on the performance of the student on labeled data (bottom arrow).}
\end{figure*}

The methods of Pseudo Labels or self-training~\cite{scudder1965probability, yarowsky1995unsupervised,riloff1996automatically,pseudo_label} have been applied successfully to improve state-of-the-art models in many computer vision tasks such as image classification (e.g.,~\cite{billion_large_scale,xie2020self}), object detection, and semantic segmentation (e.g.,~\cite{zoph2020rethinking,radosavovic2017data}).
% For example, Noisy Student~\cite{xie2020self}, an instance of Pseudo Labels methods, achieves 88.4\% top-1 accuracy on ImageNet by using an additional 300M unlabeled images.
Pseudo Labels methods work by having a pair of networks, one as a teacher and one as a student. The teacher generates pseudo labels on unlabeled images. These pseudo labeled images are then combined with labeled images to train the student. Thanks to the abundance of pseudo labeled data and the use of regularization methods such as data augmentation, the student learns to become better than the teacher~\cite{xie2020self}.

Despite the strong performance of Pseudo Labels methods, they have one main drawback: %the teacher is pre-trained, and then is fixed throughout the course of the student's training. As a result, 
if the pseudo labels are inaccurate, the student will learn from inaccurate data. As a result, the student may not get significantly better than the teacher. This drawback is also known as the problem of confirmation bias in pseudo-labeling~\cite{confirmation_bias}.
% In Pseudo Labels, the student has no mechanism to inform the teacher to change the incorrect pseudo labels.
% TODO: \textcolor{red}{use another terminology. ``confirmation bias'' means something else}

In this paper, we  design a systematic mechanism for the teacher to correct the bias by observing how its pseudo labels would affect the student.
Specifically, we propose Meta Pseudo Labels, which utilizes the feedback from the student to inform the teacher to generate better pseudo labels.
% In this paper, we propose Meta Pseudo Labels with the goal of addressing the confirmation bias in Pseudo Labels. Key to Meta Pseudo Labels is the use of feedback from the student to inform the teacher to generate better pseudo labels.
In our implementation, the feedback signal is the performance of the student on the labeled dataset. 
This feedback signal is used as a reward to train the teacher throughout the course of the student's learning. In summary, the teacher and student of Meta Pseudo Labels are trained in parallel: (1) the student learns from a minibatch of pseudo labeled data annotated by the teacher, and (2) the teacher learns from the reward signal of how well the student performs on a minibatch drawn from the labeled dataset.

% We experiment with Meta Pseudo Labels on ImageNet. We use ImageNet as labeled data, and \textcolor{red}{300M unlabeled images filtered from the JFT dataset as unlabeled data.}
We experiment with Meta Pseudo Labels, using the ImageNet~\cite{imagenet} dataset as labeled data and the JFT-300M dataset~\cite{knowledge_distillation,sun2017revisiting} as unlabeled data. We train a pair of EfficientNet-L2 networks, one as a teacher and one as a student, using Meta Pseudo Labels. The resulting student network achieves the top-1 accuracy of 90.2\% on the ImageNet ILSVRC 2012 validation set~\cite{imagenet}, which is 1.6\% better than the previous record of 88.6\%~\cite{foret2020sharpnessaware}. This student model also generalizes to the ImageNet-ReaL test set~\cite{real_paper}, as summarized in Table~\ref{tab:summary_of_results}. Small scale semi-supervised learning experiments with standard ResNet models on CIFAR-10-4K, SVHN-1K, and ImageNet-10\% also show that Meta Pseudo Labels outperforms a range of other recently proposed methods such as FixMatch~\cite{fixmatch} and Unsupervised Data Augmentation~\cite{uda}. %\textcolor{red}{Maybe add some comments in the experiments regarding how hyperparameters are selected in relation to go/imagenet-tips}

\begin{table}[H]
  \centering
  \resizebox{\linewidth}{!}{
  \begin{tabular}{lcc}
    \toprule
    \multirow{2}{*}{\textbf{Datasets}} &
    \textbf{ImageNet} & \textbf{ImageNet-ReaL} \\
    & Top-1 Accuracy & Precision@1 \\
    \midrule
    Previous SOTA~\cite{foret2020sharpnessaware,vision_transformer} &
    88.6 & 90.72 \\
    \midrule
    Ours & \textbf{90.2} & \textbf{91.02} \\
    \bottomrule
  \end{tabular}
  }
  \captionof{table}{\label{tab:summary_of_results}Summary of our key results on ImageNet ILSVRC 2012 validation set~\cite{imagenet} and the ImageNet-ReaL test set~\cite{real_paper}.}
\end{table}

%Our intuition for Meta Pseudo Labels is based on the observation that coaches in professional sports also adapt to players' performance to be more effective at designing exercises to train them. In fact, in many sports, even though coaches are not as good as players, they observe players and provide instructions to improve their performance.
% \textcolor{red}{I use metadata to connect the dots but you remove it?}
% the teacher uses metadata about the student, which is the performance on the dataset of interest, to improve the student.

% \input{methods_hieu.tex}
% \input{methods_zihang.tex}

\section{\label{sec:method}Meta Pseudo Labels}
An overview of the contrast between Pseudo Labels and Meta Pseudo Labels is presented in Figure~\ref{fig:mpl_process}. The main difference is that in Meta Pseudo Labels, the teacher receives feedback of the student's performance on a labeled dataset.

\paragraph{Notations.}
Let $T$ and $S$ respectively be the teacher network and the student network in Meta Pseudo Labels. Let their corresponding parameters be $\theta_T$ and $\theta_S$. We use $(x_l, y_l)$ to refer to a batch of images and their corresponding labels, e.g., ImageNet training images and their labels, and use $x_u$ to refer to a batch of unlabeled images, e.g., images from the internet. We denote by $T(x_u; \theta_T)$ the \textit{soft} predictions of the teacher network on the batch $x_u$ of unlabeled images and likewise for the student, e.g. $S(x_l; \theta_S)$ and $S(x_u; \theta_S)$. We use $\text{CE}(q, p)$ to denote the cross-entropy loss between two distributions $q$ and $p$; if $q$ is a label then it is understood as a one-hot distribution; if $q$ and $p$ have multiple instances in them then $\text{CE}(q, p)$ is understood as the \textit{average} of all instances in the batch. For example, $\text{CE}\big(y_l, S(x_l; \theta_S)\big)$ is the canonical cross-entropy loss in supervised learning.

\paragraph{Pseudo Labels as an optimization problem.} 
To introduce Meta Pseudo Labels, let's first review Pseudo Labels.
Specifically, Pseudo Labels (PL) trains the student model to minimize the cross-entropy loss on unlabeled data:
\begin{equation}
\label{eqn:student}
\theta_S^\text{PL} 
= \argmin_{\theta_S}\;\; \underbrace{ \mathbb{E}_{x_u} \Big[ \text{CE}\big(T(x_u; \theta_T), S(x_u; \theta_S)\big) \Big] }_{
\coloneqq \mathcal{L}_u\big(\theta_T, \theta_S\big) }
\end{equation}
where the pseudo target $T(x_u; \theta_T)$ is produced by a well pre-trained teacher model with \textit{fixed} parameter $\theta_T$.
Given a good teacher, the hope of Pseudo Labels is that the obtained $\theta_S^\text{PL}$ would ultimately achieve a low loss on labeled data, i.e. $\mathbb{E}_{x_l, y_l} \Big[ \text{CE}\big( y_l, S(x_l; \theta_S^\text{PL}) \big) \Big] \coloneqq \mathcal{L}_l\big(\theta_S^\text{PL}\big)$.

Under the framework of Pseudo Labels, notice that the optimal student parameter $\theta_S^\text{PL}$ always depends on the teacher parameter $\theta_T$ via the pseudo targets $T(x_u; \theta_T)$. 
To facilitate the discussion of Meta Pseudo Labels, we can explicitly express the dependency as $\theta_S^\text{PL}(\theta_T)$.
As an immediate observation, the ultimate student loss on labeled data $\mathcal{L}_l\big(\theta_S^\text{PL}(\theta_T)\big)$ is also a ``function'' of $\theta_T$. Therefore, we could further optimize $\mathcal{L}_l$ with respect to $\theta_T$:
\begin{equation}
\begin{aligned}
\label{eqn:bilevel_optim}
\min_{\theta_T}&\quad \mathcal{L}_l\left(\theta_S^\text{PL}(\theta_T)\right), \\
\text{where}&\quad \theta_S^\text{PL}(\theta_T) 
    = \argmin_{\theta_S} \mathcal{L}_u\big(\theta_T, \theta_S\big).
\end{aligned}
\end{equation}
Intuitively, by optimizing the teacher's parameter according to the performance of the student on labeled data, the pseudo labels can be adjusted accordingly to further improve student's performance.
As we are effectively trying to optimize the teacher on a meta level, we name our method \textit{Meta Pseudo Labels}.
However, the dependency of $\theta_S^\text{PL}(\theta_T)$ on $\theta_T$ is extremely complicated, as computing the gradient $\nabla_{\theta_T} \theta_S^\text{PL}(\theta_T)$ requires unrolling the entire student training process (i.e. $\argmin_{\theta_S}$).

\paragraph{Practical approximation.}
To make Meta Pseudo Labels feasible, we borrow ideas from previous work in meta learning~\cite{darts,maml} and approximate the multi-step $\argmin_{\theta_S}$ with the one-step gradient update of $\theta_S$:
\[
    \theta_S^\text{PL}(\theta_T) \approx \theta_S - \eta_S \cdot \nabla_{\theta_S} \mathcal{L}_u\big(\theta_T, \theta_S\big),
\]
where $\eta_S$ is the learning rate. Plugging this approximation into the optimization problem in Equation~\ref{eqn:bilevel_optim} leads to the practical teacher objective in Meta Pseudo Labels:
\begin{equation}
\label{eqn:teacher}
\min_{\theta_T}\quad \mathcal{L}_l\Big(
        \theta_S - \eta_S \cdot \nabla_{\theta_S} \mathcal{L}_u\big(\theta_T, \theta_S\big) \Big).
\end{equation}
Note that, if \textit{soft} pseudo labels are used, i.e. $T(x_u; \theta_T)$ is the full distribution predicted by teacher, the objective above is fully differentiable with respect to $\theta_T$ and we can perform standard back-propagation to get the gradient.\footnote{When optimizing Equation~\eqref{eqn:teacher}, we always treat $\theta_S$ as fixed parameters and ignore its higher-order dependency on $\theta_T$.}
However, in this work, we sample the \textit{hard} pseudo labels from the teacher distribution to train the student.
We use hard pseudo labels because they result in smaller computational graphs which are necessary for our large-scale experiments in Section~\ref{sec:large_scale_exp}.
For smaller experiments where we can use either soft pseudo labels or hard pseudo labels, we do not find significant performance difference between them.
A caveat of using hard pseudo labels is that we need to rely on a slightly modified version of REINFORCE to obtain the approximated gradient of $\mathcal{L}_l$ in Equation~\ref{eqn:teacher} with respect to $\theta_T$. We defer the detailed derivation to Appendix~\ref{sec:theta_t_grad}.

On the other hand, the student's training still relies on the objective in Equation~\ref{eqn:student}, except that the teacher parameter is \textit{not fixed} anymore. Instead, $\theta_T$ is constantly changing due to the teacher's optimization.
More interestingly, the student's parameter update can be reused in the one-step approximation of the teacher's objective, which naturally gives rise to an alternating optimization procedure between the student update and the teacher update:
\begin{itemize}[leftmargin=*]
  \item Student: draw a batch of unlabeled data $x_u$, then sample $T(x_u; \theta_T)$ from teacher's prediction, and optimize objective~\ref{eqn:student} with SGD: $\theta_S' = \theta_S - \eta_S \nabla_{\theta_S}\mathcal{L}_u(\theta_T, \theta_S)$,
  \item Teacher: draw a batch of labeled data $(x_l, y_l)$, and ``reuse'' the student's update to optimize objective~\ref{eqn:teacher} with SGD:
    $\theta_T' = \theta_T - \eta_T \nabla_{\theta_T} \mathcal{L}_l\big(\underbrace{ \theta_S - \nabla_{\theta_S} \mathcal{L}_u\big(\theta_T, \theta_S\big) }_\text{= $\theta_S'$ reused from student's update} \big)$.
\end{itemize}

\paragraph{Teacher's auxiliary losses.}
We empirically observe that Meta Pseudo Labels works well on its own. Moreover, it works even better if the teacher is jointly trained with other auxiliary objectives. Therefore, in our implementation, we augment the teacher's training with a supervised learning objective and a semi-supervised learning objective. For the supervised objective, we train the teacher on labeled data. For the semi-supervised objective, we additionally train the teacher on unlabeled data using the UDA objective~\cite{uda}. For the full pseudo code of Meta Pseudo Labels when it is combined with supervised and UDA objectives for the teacher, please see Appendix~\ref{sec:mpl_uda}, Algorithm~\ref{alg:training_procedure_uda}.

\begin{figure*}[tbh]
  \centering
  \includegraphics[width=0.82\textwidth]{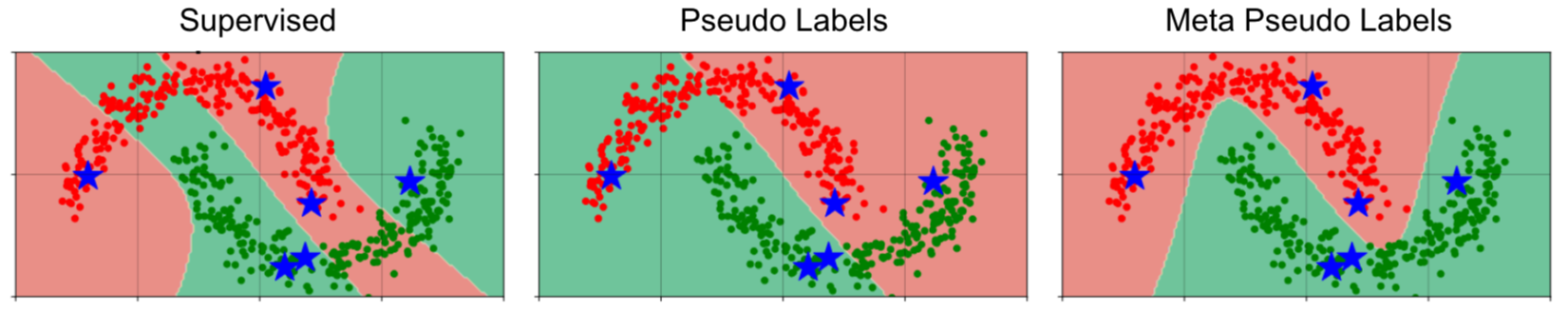}
  \caption{\label{fig:two_moons}An illustration of the importance of feedback in Meta Pseudo Labels (right). In this example, Meta Pseudo Labels works better than Supervised Learning (left) and Pseudo Labels (middle) on the simple TwoMoon dataset. More details are in Section~\ref{sec:twomoon}.
  }
\end{figure*}

Finally, as the student in Meta Pseudo Labels only learns from unlabeled data with pseudo labels generated by the teacher, we can take a student model that has converged after training with Meta Pseudo Labels and finetune it on labeled data to improve its accuracy. Details of the student's finetuning are reported in our experiments.

Next, we will present the experimental results of Meta Pseudo Labels, and organize them as follows:
\begin{itemize}[leftmargin=*]
    \item Section~\ref{sec:exp} presents small scale experiments where we compare Meta Pseudo Labels against other state-of-the-art semi-supervised learning methods on widely used benchmarks. 
    \item Section~\ref{sec:large_scale_exp} presents large scale experiments of Meta Pseudo Labels where we push the limits of ImageNet accuracy.
    %\item Control Experiments to study the importance of feedback in Meta Pseudo Labels. \textcolor{red}{Maybe we should move this section to Appendix?}
\end{itemize}

\section{\label{sec:exp}Small Scale Experiments}
In this section, we  present our empirical studies of Meta Pseudo Labels at small scales. We first study the role of feedback in Meta Pseudo Labels on the simple TwoMoon dataset~\cite{chapelle2010semi}. This study visually illustrates Meta Pseudo Labels' behaviors and benefits. We then compare Meta Pseudo Labels against state-of-the-art semi-supervised learning methods on standard benchmarks such as CIFAR-10-4K, SVHN-1K, and ImageNet-10\%. We conclude the section with experiments on the standard ResNet-50 architecture with the full ImageNet dataset.

\subsection{\label{sec:twomoon}TwoMoon Experiment}
To understand the role of feedback in Meta Pseudo Labels, we conduct an experiment on the simple and classic TwoMoon dataset~\cite{chapelle2010semi}. The 2D nature of the TwoMoon dataset allows us to visualize how Meta Pseudo Labels behaves compared to Supervised Learning and Pseudo Labels.

\paragraph{Dataset.}
For this experiment, we generate our own version of the TwoMoon dataset. In our version, there are 2,000 examples forming two clusters each with 1,000 examples. Only 6 examples are labeled, 3 examples for each cluster, while the remaining examples are unlabeled. Semi-supervised learning algorithms are asked to use these 6 labeled examples and the clustering assumption to separate the two clusters into correct classes. 

\paragraph{Training details.}
Our model architecture is a feed-forward fully-connected neural network with two hidden layers, each has 8 units. The sigmoid non-linearity is used at each layer. In Meta Pseudo Labels, both the teacher and the student share this architecture but have independent weights. All networks are trained with SGD using a constant learning rate of 0.1. The networks' weights are initialized with the uniform distribution between -0.1 and 0.1. We do not apply any regularization.

\paragraph{Results.}
We randomly generate the TwoMoon dataset for a few times and repeat the three methods: Supervised Learning, Pseudo Labels, and Meta Pseudo Labels. We observe that Meta Pseudo Labels has a much higher success rate of finding the correct classifier than Supervised Learning and Pseudo Labels. Figure~\ref{fig:two_moons} presents a typical outcome of our experiment, where the red and green regions correspond to the classifiers' decisions. As can be seen from the figure, Supervised Learning finds a bad classifier which classifies the labeled instances correctly but fails to take advantage of the clustering assumption to separate the two ``moons''. Pseudo Labels uses the bad classifier from Supervised Learning and hence receives incorrect pseudo labels on the unlabeled data. As a result, Pseudo Labels finds a classifier that misclassifies half of the data, including a few labeled instances. Meta Pseudo Labels, on the other hand, uses the feedback from the student model's loss on the labeled instances to adjust the teacher to generate better pseudo labels. As a result, Meta Pseudo Labels finds a good classifier for this dataset. In other words, Meta Pseudo Labels can address the problem of confirmation bias~\cite{confirmation_bias} of Pseudo Labels in this experiment.

\subsection{\label{sec:low_resource}CIFAR-10-4K, SVHN-1K, and ImageNet-10\% Experiments}

\paragraph{Datasets.} 
We consider three standard benchmarks: CIFAR-10-4K, SVHN-1K, and ImageNet-10\%, which have been widely used in the literature to fairly benchmark semi-supervised learning algorithms. These benchmarks were created by keeping a small fraction of the training set as labeled data while using the rest as unlabeled data. For CIFAR-10~\cite{cifar10}, 4,000 labeled examples are kept as labeled data while 41,000 examples are used as unlabeled data. The test set for CIFAR-10 is standard and consists of 10,000 examples. For SVHN~\cite{svhn}, 1,000 examples are used as labeled data whereas about 603,000 examples are used as unlabeled data. The test set for SVHN is also standard, and has 26,032 examples. Finally, for ImageNet~\cite{imagenet}, 128,000 examples are used as labeled data which is approximately 10\% of the whole ImageNet training set while the rest of 1.28 million examples are used as unlabeled data. The test set for ImageNet is the standard ILSVRC 2012 version that has 50,000 examples. We use the image resolution of 32x32 for CIFAR-10 and SVHN, and 224x224 for ImageNet.% These experimental settings are also called CIFAR-10-4K, SVHN-1K and ImageNet-10\% in the literature.

\paragraph{Training details.} 
In our experiments, our teacher and our student share the same architecture but have independent weights. For CIFAR-10-4K and SVHN-1K, we use a WideResNet-28-2~\cite{wide_res_net} which has 1.45 million parameters. For ImageNet, we use a ResNet-50~\cite{res_net} which has 25.5 million parameters. These architectures are also commonly used by previous works in this area. During the Meta Pseudo Labels training phase where we train both the teacher and the student, we use the default hyper-parameters from previous work for all our models, except for a few modifications in RandAugment~\cite{rand_augment} which we detail in Appendix~\ref{sec:random_augment}. All hyper-parameters are reported in Appendix~\ref{sec:hyper_parameters}. After training both the teacher and student with Meta Pseudo Labels, we finetune the student on the labeled dataset. For this finetuning phase, we use SGD with a fixed learning rate of $10^{-5}$ and a batch size of 512, running for 2,000 steps for ImageNet-10\% and 1,000 steps for CIFAR-10 and SVHN. Since the amount of labeled examples is limited for all three datasets, we do not use any heldout validation set. Instead, we return the model at the final checkpoint.

\paragraph{Baselines.} 
To ensure a fair comparison, we only compare Meta Pseudo Labels against methods that use the same architectures and do not compare against methods that use larger architectures such as Larger-WideResNet-28-2 and PyramidNet+ShakeDrop for CIFAR-10 and SVHN~\cite{mixmatch,berthelot2020remixmatch,wang2019enaet,uda}, or ResNet-50$\times$\{2,3,4\}, ResNet-101, ResNet-152, etc. for ImageNet-10\%~\cite{cpc_v2,moco,moco_v2,simclr,simclrv2}. We also do not compare Meta Pseudo Labels with training procedures that include self-distillation or distillation from a larger teacher~\cite{simclr,simclrv2}. We enforce these restrictions on our baselines since it is known that larger architectures and distillation can improve any method, possibly including Meta Pseudo Labels. % \textcolor{red}{any model, but such improvements should not be associated to a training algorithm's merit. Hieu: should we add this back?} %\textcolor{red}{i think we need this paragraph. what do you think? Seems good to me.}

We directly compare Meta Pseudo Labels against two baselines: Supervised Learning with full dataset and Unsupervised Data Augmentation (UDA~\cite{uda}). Supervised Learning with full dataset represents the headroom because it unfairly makes use of all labeled data (e.g., for CIFAR-10, it uses all 50,000 labeled examples). We also compare against UDA because our implementation of Meta Pseudo Labels uses UDA in training the teacher. Both of these baselines use the same experimental protocols and hence ensure a fair comparison. We follow~\cite{realistic_eval}'s train/eval/test splitting, and we use the same amount of resources to tune hyper-parameters for our baselines as well as for Meta Pseudo Labels. More details are in Appendix~\ref{sec:exp_details}.

\begin{table*}[h!]
\small
\centering
\resizebox{0.92\linewidth}{!}{ %
\begin{tabular}{llcccc}
  \toprule
  & \multirow{2}{*}{\textbf{Method}} & \textbf{CIFAR-10-4K} & \textbf{SVHN-1K} & \multicolumn{2}{c}{\textbf{ImageNet-10\%}} \\
  & & (mean $\pm$ std) & (mean $\pm$ std) & Top-1 & Top-5 \\
  \midrule
  \multirow{10}{*}{Label Propagation Methods}
  & Temporal Ensemble~\cite{temporal_ensemble} & $83.63 \pm 0.63$ & $92.81 \pm 0.27$ & \multicolumn{2}{c}{$-$} \\
  & Mean Teacher~\cite{mean_teacher} & $84.13 \pm 0.28$ & $94.35 \pm 0.47$ & \multicolumn{2}{c}{$-$} \\
  & VAT + EntMin~\cite{vat} & $86.87 \pm 0.39$ & $94.65 \pm 0.19$ & $-$ & 83.39 \\
  & LGA + VAT~\cite{lga} & $87.94 \pm 0.19$ & $93.42 \pm 0.36$ & \multicolumn{2}{c}{$-$} \\
  & ICT~\cite{ict} & $92.71 \pm 0.02$ & $96.11 \pm 0.04$ & \multicolumn{2}{c}{$-$} \\
  & MixMatch~\cite{mixmatch} & $93.76 \pm 0.06$ & $96.73 \pm 0.31$ & \multicolumn{2}{c}{$-$} \\
  & ReMixMatch~\cite{berthelot2020remixmatch} & $94.86 \pm 0.04$ & $97.17 \pm 0.30$ & \multicolumn{2}{c}{$-$} \\
  & EnAET~\cite{wang2019enaet} & $94.65$ & $97.08$ & \multicolumn{2}{c}{$-$} \\
  & FixMatch~\cite{fixmatch} & $95.74 \pm  0.05$ & $97.72 \pm 0.38$ & $71.5$& $89.1$ \\
  & UDA$^*$~\cite{uda}& $94.53 \pm 0.18$ & $97.11 \pm 0.17$ & $68.07$ & $88.19$ \\
  \midrule
  \multirow{5}{*}{Self-Supervised Methods}
  & SimCLR~\cite{simclr,simclrv2} & $-$ & $-$ & $71.7$  &$90.4$\\
  & MOCOv2~\cite{moco_v2} & $-$ & $-$ & $71.1$ & $-$ \\
  & PCL~\cite{li2020prototypical} & $-$ & $-$ & $-$ & $85.6$ \\
  & PIRL~\cite{misra2020self} & $-$ & $-$ & $-$ & $84.9$ \\
  & BYOL~\cite{grill2020bootstrap} & $-$ & $-$ & $68.8$ & $89.0$ \\
  \midrule
  & \bf Meta Pseudo Labels & \textbf{96.11 $\pm$ 0.07} & \textbf{98.01 $\pm$ 0.07} & \textbf{73.89} & \textbf{91.38} \\
  \cmidrule{2-6}
  & Supervised Learning with full dataset$^*$ & $94.92 \pm 0.17$ & $97.41 \pm 0.16$ & $76.89$  &$93.27$ \\
  \bottomrule
\end{tabular}
} %
\caption{\label{tab:low_resource_results}Image classification accuracy on CIFAR-10-4K, SVHN-1K, and ImageNet-10\%. Higher is better. For CIFAR-10-4K and SVHN-1K, we report $\text{mean}\pm\text{std}$ over 10 runs, while for ImageNet-10\%, we report $\text{Top-1}/\text{Top-5}$ accuracy of a single run. For fair comparison, we only include results that share the same model architecture: WideResNet-28-2 for CIFAR-10-4K and SVHN-1K, and ResNet-50 for ImageNet-10\%. $^*$ indicates our implementation which uses the same experimental protocols. Except for UDA, results in the first two blocks are from representative important papers, and hence do not share the same controlled environment with ours.}
\end{table*}

\paragraph{Additional baselines.} 
In addition to these two baselines, we also include a range of other semi-supervised baselines in two categories: Label Propagation and Self-Supervised. Since these methods do not share the same controlled environment, the comparison to them is not direct, and should be contextualized as suggested by~\cite{realistic_eval}. More controlled experiments comparing Meta Pseudo Labels to other baselines are presented in Appendix~\ref{sec:exp_analysis}. % \textcolor{red}{Check if the Appendix indeed contains the details.}

\paragraph{Results.} 
Table~\ref{tab:low_resource_results} presents our results with Meta Pseudo Labels in comparison with other methods. The results show that under strictly fair comparisons (as argued by~\cite{realistic_eval}), Meta Pseudo Labels significantly improves over UDA. Interestingly, on CIFAR-10-4K, Meta Pseudo Labels even exceeds the headroom supervised learning on full dataset. On ImageNet-10\%, Meta Pseudo Labels outperforms the UDA teacher by more than 5\% in top-1 accuracy, going from 68.07\% to 73.89\%. For ImageNet, such relative improvement is very significant.

\paragraph{Comparing to existing state-of-the-art methods.} 
Compared to results reported from past papers, Meta Pseudo Labels has achieved the best accuracies \textit{among the same model architectures} on all the three datasets: CIFAR-10-4K, SVHN-1K, and ImageNet-10\%. On CIFAR-10-4K and SVHN-1K, Meta Pseudo Labels leads to almost 10\% relative error reduction compared to the highest reported baselines~\cite{fixmatch}. On ImageNet-10\%, Meta Pseudo Labels outperforms SimCLR~\cite{simclr,simclrv2} by 2.19\% top-1 accuracy. 

While better results on these datasets exist, to our knowledge, such results are all obtained with larger models, stronger regularization techniques, or extra distillation procedures.
For example, the best reported accuracy on CIFAR-10-4K is 97.3\%~\cite{uda} but this accuracy is achieved with a PyramidNet which has 17x more parameters than our WideResNet-28-2 and uses the complex ShakeDrop regularization~\cite{shake_drop}.
%and which is trained using ShakeDrop regularization~\cite{shake_drop}.
On the other hand, the best reported top-1 accuracy for ImageNet-10\% is 80.9\%, achieved by SimCLRv2~\cite{simclrv2} using a self-distillation training phase and a ResNet-152$\times$3 which has 32x more parameters than our ResNet-50.
Such enhancements on architectures, regularization, and distillation can also be applied to Meta Pseudo Labels to further improve our results.

\subsection{\label{sec:resnet50}ResNet-50 Experiment}
%We have seen Meta Pseudo Labels achieves strong performance for low data image classification tasks. Another aspect of these tasks is that the unlabeled data also come from the same domain as the labeled data, which is a restricted assumption. In this section, we show that Meta Pseudo Labels also excels in the regime where we have a large labeled dataset and an order of magnitude more unlabeled data. In this regime, we also test the performance of our method when the unlabeled set may have out-of-domain images, \ie~the images belong to categories that do not exist in ImageNet.

The previous experiments show that Meta Pseudo Labels outperforms other semi-supervised learning methods on CIFAR-10-4K, SVHN-1K, and ImageNet-10\%. In this experiment, we benchmark Meta Pseudo Labels on the entire ImageNet dataset plus unlabeled images from the JFT dataset. The purpose of this experiment is to verify if Meta Pseudo Labels works well on the widely used ResNet-50 architecture~\cite{res_net} before we conduct more large scale experiments on EfficientNet (Section~\ref{sec:large_scale_exp}).

\paragraph{Datasets.} 
As mentioned, we experiment with all labeled examples from the ImageNet dataset. We reserve 25,000 examples from the ImageNet dataset for hyper-parameter tuning and model selection. Our test set is the ILSVRC 2012 validation set. Additionally, we take 12.8 million unlabeled images from the JFT dataset. To obtain these 12.8 million unlabeled images, we first train a ResNet-50 on the entire ImageNet training set and then use the resulting ResNet-50 to assign class probabilities to images in the JFT dataset. We then select 12,800 images of highest probability for each of the 1,000 classes of ImageNet. This selection results in 12.8 million images. We also make sure that none of the 12.8 million images that we use overlaps with the ILSVRC 2012 validation set of ImageNet.  This procedure of filtering extra unlabeled data has been used by UDA~\cite{uda} and Noisy Student~\cite{xie2020self}.

\paragraph{Implementation details.} 
We implement Meta Pseudo Labels the same as in Section~\ref{sec:low_resource} but we use a larger batch size and more training steps, as the datasets are much larger for this experiment. Specifically, for both the student and the teacher, we use the batch size of 4,096 for labeled images and the batch size of 32,768 for unlabeled images.
We train for 500,000 steps which equals to about 160 epochs on the unlabeled dataset. After training the Meta Pseudo Labels phase on ImageNet+JFT, we finetune the resulting student on ImageNet for 10,000 SGD steps, using a fixed learning rate of $10^{-4}$. Using 512 TPUv2 cores, our training procedure takes about 2 days.

% , except for one minor detail: Instead of directly training and consistently regularizing the teacher, we initialize the teacher using a pre-trained ResNet-50 (pre-trained on full ImageNet). Then, throughout the course of the student's learning, we only train the teacher to minimize the student's cross-entropy loss. We do not use additional supervised loss for the teacher because  once the teacher is pre-trained, adding another loss to the teacher has minimal effect. We do not consistently regularize the teacher because~\cite{uda} has found that consistency regularization requires in-domain data, while we do not filter our unlabeled images from OpenImages. \textcolor{red}{Some of these details are a bit complicated?}

\paragraph{Baselines.} 
We compare Meta Pseudo Labels against two groups of baselines. The first group contains supervised learning methods with data augmentation or regularization methods such as AutoAugment~\cite{auto_augment}, DropBlock\cite{drop_block}, and CutMix~\cite{cut_mix}. These baselines represent state-of-the-art supervised learning methods on ResNet-50. The second group of baselines consists of three recent semi-supervised learning methods that leverage the labeled training images from ImageNet and unlabeled images elsewhere. Specifically, billion-scale semi-supervised learning~\cite{billion_large_scale} uses unlabeled data from the YFCC100M dataset~\cite{yfcc100m}, while UDA~\cite{uda} and Noisy Student~\cite{xie2020self} both use JFT as unlabeled data like Meta Pseudo Labels. Similar to Section~\ref{sec:low_resource}, we only compare Meta Pseudo Labels to results that are obtained with ResNet-50 and without distillation.

\begin{table}[htb!]
  \small
  \centering
  \resizebox{0.92\linewidth}{!}{ %
  \begin{tabular}{lcc}
    \toprule
    \multirow{2}{*}{\textbf{Method}} & 
    \textbf{Unlabeled} &
    \textbf{Accuracy} \\
    & \textbf{Images}& (top-1$/$top-5) \\
    \midrule
    Supervised~\cite{res_net} & None & 76.9$/$93.3 \\
    AutoAugment~\cite{auto_augment} & None & 77.6$/$93.8 \\
    DropBlock~\cite{drop_block} & None & 78.4$/$94.2 \\
    FixRes~\cite{train_test_resolution} & None & 79.1$/$94.6 \\
    FixRes+CutMix~\cite{cut_mix} & None & 79.8$/$94.9 \\
    \midrule
    NoisyStudent~\cite{xie2020self} & JFT & 78.9$/$94.3 \\
    UDA~\cite{uda} & JFT & 79.0$/$94.5 \\
    Billion-scale SSL~\cite{train_test_resolution,billion_large_scale} & YFCC &  82.5$/$\textbf{96.6} \\
    \midrule
    \bf Meta Pseudo Labels & JFT & \textbf{83.2}$/$96.5 \\
    \bottomrule
  \end{tabular}
  } %
  \caption{\label{tab:resnet50_results}Top-1 and Top-5 accuracy of Meta Pseudo Labels and other representative supervised and semi-supervised methods on ImageNet with ResNet-50.}
\end{table}

\paragraph{Results.} 
Table~\ref{tab:resnet50_results} presents the results. As can be seen from the table, Meta Pseudo Labels boosts the top-1 accuracy of ResNet-50 from 76.9\% to 83.2\%, which is a large margin of improvement for ImageNet, outperforming both UDA and Noisy Student. Meta Pseudo Labels also outperforms Billion-scale SSL~\cite{train_test_resolution,billion_large_scale} in top-1 accuracy. This is particularly impressive since Billion-scale SSL pre-trains their ResNet-50 on weakly-supervised images from Instagram.

\section{\label{sec:large_scale_exp}Large Scale Experiment: Pushing the Limits of ImageNet Accuracy}

\begin{table*}[h!]
\centering
\resizebox{0.96\linewidth}{!}{ %
\begin{tabular}{lccllc}
  \toprule
  \multirow{2}{*}{\textbf{Method}} &
  \multirow{2}{*}{\textbf{\# Params}} &
  \multirow{2}{*}{\textbf{Extra Data}} &
  \multicolumn{2}{c}{\textbf{ImageNet}} &
  \textbf{ImageNet-ReaL}~\cite{real_paper} \\
  &&& Top-1 & Top-5 & Precision@1 \\
  \midrule
  ResNet-50 \cite{res_net} & 26M & $-$ & 76.0 & 93.0  & 82.94 \\
  ResNet-152 \cite{res_net} & 60M  & $-$ &  77.8 & 93.8 & 84.79 \\
  DenseNet-264 \cite{dense_net} & 34M & $-$ & 77.9 & 93.9 & $-$ \\
  Inception-v3 \cite{inception}& 24M  & $-$ & 78.8 & 94.4 & 83.58 \\
  Xception \cite{xception} & 23M & $-$ & 79.0 & 94.5 & $-$ \\
  Inception-v4 \cite{szegedy2017inception} & 48M  & $-$ & 80.0 & 95.0 & $-$ \\
  Inception-resnet-v2  \cite{szegedy2017inception} & 56M & $-$ & 80.1 & 95.1 & $-$ \\
  ResNeXt-101 \cite{resnext}  & 84M   & $-$ & 80.9 & 95.6 & 85.18 \\
  PolyNet \cite{zhang2017polynet}  & 92M  & $-$ & 81.3  & 95.8 & $-$ \\
  SENet \cite{hu2018squeeze} & 146M & $-$ & 82.7 & 96.2 & $-$ \\
  NASNet-A \cite{zoph2018learning} & 89M & $-$ & 82.7 & 96.2  & 82.56 \\
  AmoebaNet-A \cite{real2019regularized} & 87M & $-$ & 82.8 & 96.1 & $-$ \\
  PNASNet \cite{liu2018progressive} & 86M  & $-$ & 82.9 & 96.2 & $-$ \\
  AmoebaNet-C + AutoAugment \cite{auto_augment}  & 155M  & $-$ &  83.5 & 96.5 & $-$ \\
  GPipe \cite{gpipe18} & 557M & $-$ & 84.3  & 97.0 & $-$ \\
  EfficientNet-B7 \cite{efficient_net} &  66M  & $-$ & 85.0 & 97.2 & $-$ \\
  EfficientNet-B7 + FixRes \cite{touvron2020fixing} &  66M  & $-$ & 85.3 & 97.4 & $-$ \\
  EfficientNet-L2 \cite{efficient_net} &  480M  & $-$ & 85.5 & 97.5 & $-$ \\
  \midrule
  ResNet-50 Billion-scale SSL \cite{billion_large_scale} & 26M & 3.5B labeled Instagram & 81.2 & 96.0 & $-$ \\
  ResNeXt-101 Billion-scale SSL \cite{billion_large_scale} & 193M & 3.5B labeled Instagram & 84.8 & $-$ & $-$ \\
  ResNeXt-101 WSL \cite{instagram_imagenet} & 829M & 3.5B labeled Instagram & 85.4 & 97.6 & 88.19\\
  FixRes ResNeXt-101 WSL \cite{touvron2019fixing} & 829M & 3.5B labeled Instagram & 86.4 &  98.0 & 89.73 \\
  %\midrule
  Big Transfer (BiT-L) \cite{big_transfer} & 928M  & 300M  labeled  JFT & 87.5 &  98.5  & 90.54 \\
  %\midrule
  Noisy Student (EfficientNet-L2)~\cite{xie2020self} &  480M & 300M unlabeled  JFT & 88.4 & 98.7  & 90.55\\
  Noisy Student + FixRes~\cite{touvron2020fixing} & 480M & 300M unlabeled  JFT & 88.5 & 98.7 & $-$  \\
  Vision Transformer (ViT-H)~\cite{vision_transformer} & 632M & 300M labeled  JFT & 88.55 & $-$ & 90.72\\
  EfficientNet-L2-NoisyStudent + SAM~\cite{foret2020sharpnessaware} & 480M & 300M unlabeled JFT & 88.6 & 98.6 & $-$\\
  \midrule
  Meta Pseudo Labels (EfficientNet-B6-Wide) & 390M & 300M unlabeled  JFT & 90.0 & 98.7 & \textbf{91.12} \\
  Meta Pseudo Labels (EfficientNet-L2) & 480M & 300M unlabeled  JFT & \textbf{90.2} & \textbf{98.8} & 91.02 \\
  \bottomrule
\end{tabular}
} %
\caption{\label{tab:imagenet}Top-1 and Top-5 accuracy of Meta Pseudo Labels and previous state-of-the-art methods on ImageNet. With EfficientNet-L2 and EfficientNet-B6-Wide, Meta Pseudo Labels achieves an improvement of 1.6\% on top of the state-of-the-art~\cite{foret2020sharpnessaware}, despite the fact that the latter uses 300 million \textit{labeled} training examples from JFT.}
\end{table*}

In this section, we scale up Meta Pseudo Labels to train on a large model and a large dataset to push the limits of ImageNet accuracy. Specifically, we use the EfficientNet-L2 architecture because it has a higher capacity than ResNets. EfficientNet-L2 was also used by Noisy Student~\cite{xie2020self} to achieve the top-1 accuracy of 88.4\% on ImageNet. %This result is further improved by FixRes of 88.5\%~\cite{touvron2020fixing}, and  recently it is slightly outperformed by Vision Transformer of 88.55\%~\cite{vision_transformer}. %To date, this is still the best accuracy on the standard ImageNet 2012 validation dataset.

\paragraph{Datasets.} For this experiment, we use the entire ImageNet training set as labeled data, and use the JFT dataset as unlabeled data. The JFT dataset has 300 million images, and then is filtered down to 130 million images by Noisy Student using confidence thresholds and up-sampling~\cite{xie2020self}. We use the same 130 million images as Noisy Student.% The unlabeled dataset for this experiment is about 10x larger than that for our ResNet-50 (Section~\ref{sec:resnet50}). This benefits our training since EfficientNet-L2 has a much larger learning capacity than ResNet-50.

\paragraph{Model architecture.} We experiment with EfficientNet-L2 since it has the state-of-the-art performance on ImageNet~\cite{xie2020self} without extra labeled data. We use the same hyper-parameters with Noisy Student, except that we use the training image resolution of 512x512 instead of 475x475. We increase the input image resolution to be compatible with our model parallelism implementation which we discuss in the next paragraph. In addition to EfficientNet-L2, we also experiment with a smaller model, which has the same depth with EfficientNet-B6~\cite{efficient_net} but has the width factor increased from 2.1 to 5.0. This model, termed EfficientNet-B6-Wide, has 390 million parameters. We adopt all hyper-parameters of EfficientNet-L2 for EfficientNet-B6-Wide. We find that EfficientNet-B6-Wide has almost the same performance with EfficientNet-L2, but is faster to compile and train.

\paragraph{Model parallelism.} Due to the memory footprint of our networks, keeping two such networks in memory for the teacher and the student would vastly exceed the available memory of our accelerators. We thus design a hybrid model-data parallelism framework to run Meta Pseudo Labels. Specifically, our training process runs on a cluster of 2,048 TPUv3 cores. We divide these cores into 128 identical replicas to run with standard data parallelism with synchronized gradients. Within each replica, which runs on 2,048/128=16 cores, we implement two types of model parallelism. First, each input image of resolution 512x512 is split along the width dimension into 16 patches of equal size 512x32 and is distributed to 16 cores to process. Note that we choose the input resolution of 512x512 because 512 is close to the resolution 475x475 used by Noisy Student and 512 keeps the dimensions of the network's intermediate outputs divisible by 16. Second, each weight tensor is also split equally into 16 parts that are assigned to the 16 cores. We implement our hybrid data-model parallelism in the XLA-Sharding framework~\cite{gshard}. With this parallelism, we can fit a batch size of 2,048 labeled images and 16,384 unlabeled images into each training step. We train the model for 1 million steps in total, which takes about 11 days for EfficientNet-L2 and 10 days for EfficientNet-B6-Wide. After finishing the Meta Pseudo Labels training phase, we finetune the models on our labeled dataset for 20,000 steps. Details of the finetuning procedures are in Appendix~\ref{sec:hyper_parameters}.

\paragraph{Results.} Our results are presented in Table~\ref{tab:imagenet}. From the table, it can be seen that Meta Pseudo Labels achieves 90.2\% top-1 accuracy on ImageNet, which is a new state-of-the-art on this dataset. This result is 1.8\% better than the same EfficientNet-L2 architecture trained with Noisy Student~\cite{xie2020self} and FixRes~\cite{touvron2019fixing,touvron2020fixing}.
Meta Pseudo Labels also outperforms the recent results by BiT-L~\cite{big_transfer} and the previous state-of-the-art by Vision Transformer~\cite{vision_transformer}.
The important contrast here is that both Bit-L and Vision Transformer pre-train on 300 million \textit{labeled} images from JFT, while our method only uses \textit{unlabeled} images from this dataset.
At this level of accuracy, our gain of 1.6\% over~\cite{foret2020sharpnessaware} is a very significant margin of improvement compared to recent gains. For instance, the gain of Vision Transformer~\cite{vision_transformer} over Noisy Student + FixRes was only 0.05\%, and the gain of FixRes over Noisy Student was only 0.1\%.

Finally, to verify that our model does not simply overfit to the ImageNet ILSVRC 2012 validation set, we test it on the ImageNet-ReaL test set~\cite{real_paper}. On this test set, our model also works well and achieves 91.02\% Precision@1  which is 0.4\%  better than Vision Transformer~\cite{vision_transformer}. This gap is also bigger than the gap between Vision Transformer and Noisy Student which is only 0.17\%.

\paragraph{A lite version of Meta Pseudo Labels.} Given the expensive training cost of Meta Pseudo Labels, we design a lite version of Meta Pseudo Labels, termed \textit{Reduced Meta Pseudo Labels}. We describe this lite version in Appendix~\ref{sec:reduced_mpl}, where we achieve 86.9\% top-1 accuracy on the ImageNet ILSRVC 2012 validation set with EfficentNet-B7. To avoid using proprietary data like JFT, we use the ImageNet training set as labeled data and the YFCC100M dataset~\cite{yfcc100m} as unlabeled data. Reduced Meta Pseudo Labels allows us to implement the feedback mechanism of Meta Pseudo Labels while avoiding the need to keep two networks in memory. %

\section{\label{sec:related}Related Works}
%Meta Pseudo Labels are inspired by recent successes of Pseudo Labels and other related semi-supervised learning methods. The adjustment of soft labels is also related to knowledge distillation and label smoothing. The bi-level optimization in Meta Pseudo Labels is related to works in meta learning. 

\paragraph{Pseudo Labels.}
The method of Pseudo Labels, also known as self-training, is a simple Semi-Supervised Learning (SSL) approach that has been successfully applied to improve the state-of-the-art of many tasks, such as: image classification~\cite{billion_large_scale,xie2020self}, object detection, semantic segmentation~\cite{zoph2020rethinking}, machine translation~\cite{he2020revisiting}, and speech recognition~\cite{kahn2020self,park2020improved}. Vanilla Pseudo Labels methods keep a pre-trained teacher fixed during the student's learning, leading to a confirmation bias~\cite{confirmation_bias} when the pseudo labels are inaccurate. Unlike vanilla Pseudo Labels, Meta Pseudo Labels continues to adapt the teacher to improve the student's performance on a labeled dataset. This extra adaptation allows the teacher to generate better pseudo labels to teach the student as shown in our experiments.

\paragraph{Other SSL approaches.} 
Other typical SSL methods often train a single model by optimizing an objective function that combines a supervised loss on labeled data and an unsupervised loss on unlabeled data. The supervised loss is often the cross-entropy computed on the labeled data. Meanwhile, the unsupervised loss is typically either a self-supervised loss or a label propagation loss. Self-supervised losses typically encourage the model to develop a common sense about images, such as in-painting~\cite{inpainting}, solving jigsaw puzzles~\cite{jigsaw_puzzle}, predicting the rotation angle~\cite{predicting_rotation}, contrastive prediction~\cite{cpc_v2,moco_v2,simclr,simclrv2,li2020prototypical}, or bootstraping the latent space~\cite{grill2020bootstrap}. On the other hand, label propagation losses typically enforce that the model is invariant against certain transformations of the data such as data augmentations, adversarial attacks, or proximity in the latent space~\cite{temporal_ensemble,mean_teacher,vat,mixmatch,uda,lga,ict,fixmatch,ke2019dual,radosavovic2017data,grandvalet2005semi}. Meta Pseudo Labels is distinct from the aforementioned SSL methods in two notable ways. First, the student in Meta Pseudo Labels never learns directly from labeled data, which helps to avoid overfitting, especially when labeled data is limited. Second, the signal that the teacher in Meta Pseudo Labels receives from the student's performance on labeled data is a novel way of utilizing labeled data.

\paragraph{Knowledge Distillation and Label Smoothing.} 
The teacher in Meta Pseudo Labels uses its softmax predictions on unlabeled data to teach the student. These softmax predictions are generally called the soft labels, which have been widely utilized in the literature on knowledge distillation~\cite{knowledge_distillation,born_again_networks,zhang2019be}. Outside the line of work on distillation, manually designed soft labels, such as label smoothing~\cite{label_smoothing_investigation} and temperature sharpening or dampening~\cite{uda,xie2020self}, have also been shown to improve models' generalization. Both of these methods can be seen as adjusting the labels of the training examples to improve optimization and generalization. Similar to other SSL methods, these adjustments do not receive any feedback from the student's performance as proposed in this paper. An experiment comparing Meta Pseudo Labels to Label Smoothing is presented in Appendix~\ref{sec:mpl_reg}.

\paragraph{Bi-level optimization algorithms.} 
We use \emph{Meta} in our method name because our technique of deriving the teacher's update rule from the student's feedback is based on a bi-level optimization problem which appears frequently in the literature of meta-learning. Similar bi-level optimization problems have been proposed to optimize a model's learning process, such as learning the learning rate schedule~\cite{hyper_gradient_descent}, designing architectures~\cite{darts}, correcting wrong training labels~\cite{zheng2019meta}, generating training examples~\cite{such2019generative}, and re-weighting training data~\cite{wang2020optimizing,wang2020meta,ren2020not,ren2018learning}. Meta Pseudo Labels uses the same bi-level optimization technique in this line of work to derive the teacher's gradient from the student's feedback. The difference between Meta Pseudo Labels and these methods is that Meta Pseudo Labels applies the bi-level optimization technique to improve the pseudo labels generated by the teacher model.

\section{\label{sec:conclusion}Conclusion}
In this paper, we proposed the Meta Pseudo Labels method for semi-supervised learning. Key to Meta Pseudo Labels is the idea that the teacher learns from the student's feedback to generate pseudo labels in a way that best helps student's learning.
% We describe a practical method that trains both the teacher in parallel with the student.
The learning process in Meta Pseudo Labels consists of two main updates: updating the student based on the pseudo labeled data produced by the teacher and updating the teacher based on the student's performance.
Experiments on standard low-resource benchmarks such as CIFAR-10-4K, SVHN-1K, and ImageNet-10\% show that Meta Pseudo Labels is better than many existing semi-supervised learning methods. Meta Pseudo Labels also scales well to large problems,
% and successfully uses out-of-domain data to improve ImageNet classification. In particular, Meta Pseudo Labels
attaining 90.2\% top-1 accuracy on ImageNet, which is 1.6\% better than the previous state-of-the-art~\cite{foret2020sharpnessaware}. The consistent gains confirm the benefit of the student's feedback to the teacher.

\section*{Acknowledgements}
The authors wish to thank Rohan Anil, Frank Chen, Wang Tao for their help with many technical issues in running our experiments. We also thank David Berthelot, Nicholas Carlini, Sylvain Gelly, Geoff Hinton, Mohammad Norouzi, and Colin Raffel for their comments on earlier drafts of the paper, and others in the Google Brain Team for their support throughout this very long project.

Jaime Carbonell has also advised us on removing the data loading bottleneck for the ResNets model ImageNet. His advice helped a lot when we did not have enough spare TPUs for our ResNet jobs. He will be deeply remembered.

{\small
\bibliographystyle{ieee_fullname}
\bibliography{main}
}

\newpage
\appendix
\onecolumn

\section{\label{sec:theta_t_grad}Derivation of the Teacher's Update Rule}
In this section, we present the detailed derivation of the teacher's update rule in Section~\ref{sec:method}.

\paragraph{Mathematical Notations and Conventions.} Since we will work with the chain rule, we use the standard Jacobian notations.\footnote{Standard:~\url{https://en.wikipedia.org/wiki/Jacobian_matrix_and_determinant}} Specifically, for a differentiable function $f: \mathbb{R}^{m} \to \mathbb{R}^{n}$, and for a vector $x \in \mathbb{R}^{m}$, we use the notation $\frac{\partial f}{\partial x} \in \mathbb{R}^{n \times m}$ to denote \textit{the Jacobian matrix} of $f$, whose dimension is $n \times m$. Additionally, when we mention the Jacobian of a function $f$ at multiple points such as $x_1$ and $x_2$, we will use the notations of $\left. \frac{\partial f}{\partial x} \right|_{x = x_1}$ and $\left. \frac{\partial f}{\partial x} \right|_{x = x_2}$.

Furthermore, by mathematical conventions, a vector $v \in \mathbb{R}^{n}$ is treated as a \textit{column matrix} -- that is, a matrix of size $n \times 1$. For this reason, the gradient vector of a multi-variable real-valued function is actually the transpose of of its Jacobian matrix. Finally, all multiplications in this section are standard matrix multiplications. If an operand is a vector, then the operand is treated as a column matrix.

\paragraph{Dimension Annotations.} Understanding that these notations and conventions might cause confusions, in the derivation below, we annotate the dimensions of the computed quantities to ensure that there is no confusion caused to our readers. To this end, we respectively use $\ABS{S}$ and $\ABS{T}$ to denote the dimensions of the parameters $\theta_S$, $\theta_T$. That is, $\theta_S \in \mathbb{R}^{\ABS{S} \times 1}$ and $\theta_T \in \mathbb{R}^{\ABS{T} \times 1}$.

We now present the derivation. Suppose that on a batch of unlabeled examples $x_{u}$, the teacher samples the pseudo labels $\widehat{y}_u \sim T(x_u; \theta_T)$ and the student uses $(x_u, \widehat{y}_{u})$ to update its parameter $\theta_S$. In expectation, the student's new parameter is $\mathbb{E}_{\widehat{y}_{u} \sim T(x_u; \theta_T)}\big[ \theta_S - \eta_S \nabla_{\eta_S} \text{CE} (\widehat{y}_{u}, S(x_{u}; \theta_S)) \big]$. We will update the teacher's parameter to minimize the student's cross-entropy on a batch of labeled data a this expected parameter. To this end, we need to compute the Jacobian:
\begin{equation}
  \begin{aligned}
    \underbrace{\frac{\partial R}{\partial \theta_T}}_{1 \times \ABS{T}}
    = \frac{\partial}{\partial \theta_T} \text{CE} \left(
      y_{l},
      S\Big (x_{l}; \mathbb{E}_{\widehat{y}_{u} \sim T(x_u; \theta_T)}\big[ \theta_S - \eta_S \nabla_{\eta_S} \text{CE} (\widehat{y}_{u}, S(x_{u}; \theta_S)) \big] \Big)
    \right)
  \end{aligned}
\end{equation}
To simplify our notation, let us define
\begin{equation}
  \begin{aligned}
    \underbrace{\bar{\theta}_S'}_{\ABS{S} \times 1}
    = \mathbb{E}_{\widehat{y}_{u} \sim T(x_u; \theta_T)}\big[ \theta_S - \eta_S \nabla_{\eta_S} \text{CE} (\widehat{y}_{u}, S(x_{u}; \theta_S)) \big]
  \end{aligned}
\end{equation}
Then, by the chain rule, we have
\begin{equation}
  \label{eqn:theta_t_chain_rule}
  \begin{aligned}
    \underbrace{\frac{\partial R}{\partial \theta_T}}_{1 \times \ABS{T}}
    &= \frac{\partial}{\partial \theta_T} \text{CE} \left(
      y_{l},
      S\Big (x_{l}; \mathbb{E}_{\widehat{y}_{u} \sim T(x_u; \theta_T)}\big[ \theta_S - \eta_S \nabla_{\eta_S} \text{CE} (\widehat{y}_{u}, S(x_{u}; \theta_S)) \big] \Big)
    \right) \\
    &= \frac{\partial}{\partial \theta_T} \text{CE} \left(
      y_{l}, S\big(x_{l}; \bar{\theta}_S' \big)
    \right) \\
    &= \underbrace{\left. \frac{\partial \text{CE} \left(
      y_{l}, S\big (x_{l}; \bar{\theta}_S' \big) \right) }{\partial \theta_S} \right|_{\theta_S = \bar{\theta}_S' \big) }}_{1 \times \ABS{S}}
    \cdot
    \underbrace{\frac{\partial \bar{\theta}_S'}{\partial \theta_T}}_{\ABS{S} \times \ABS{T}}
  \end{aligned}
\end{equation}
The first factor in Equation~\ref{eqn:theta_t_chain_rule} can be simply computed via back-propagation. We now focus on the second term. We have
\begin{equation}
  \label{eqn:theta_t_second_term}
  \begin{aligned}
    \underbrace{\frac{\partial \bar{\theta}_S' }{\partial \theta_T}}_{\ABS{S} \times \ABS{T}}
    &= \frac{\partial}{\partial \theta_T} \mathbb{E}_{\widehat{y}_{u} \sim T(x_u; \theta_T)}\big[ \theta_S - \eta_S \nabla_{\eta_S} \text{CE} (\widehat{y}_{u}, S(x_{u}; \theta_S)) \big] \\
    &= \frac{\partial}{\partial \theta_T} \mathbb{E}_{\widehat{y}_{u} \sim T(x_u; \theta_T)} \left[ \theta_S - \eta_S \cdot \left( \left. \frac{\partial \text{CE} \left( \widehat{y}_{u}, S(x_{u}; \theta_S) \right)}{\partial \theta_S} \right|_{\theta_S = \theta_S} \right)^\top \right]
  \end{aligned}
\end{equation}
Note that in Equation~\ref{eqn:theta_t_second_term} above, the Jacobian of $\text{CE} \left( \widehat{y}_{u}, S(x_{u}; \theta_S) \right)$, which has dimension $1 \times \ABS{S}$, needs to be transposed to match the dimension of $\theta_S$, which, as we discussed above, conventionally has dimension $\ABS{S} \times 1$.

Now, since $\theta_S$ in Equation~\ref{eqn:theta_t_second_term} does not depend on $\theta_T$, we can leave it out of subsequent derivations. Also, to simplify notations, let us define \textit{the gradient}
\begin{equation}
  \label{eqn:g_S_t_expected}
  \begin{aligned}
    \underbrace{g_S(\widehat{y}_{u})}_{\ABS{S} \times \ABS{1}}
    =
    \left( \left. \frac{\partial \text{CE} \left( \widehat{y}_{u}, S(x_{u}; \theta_S) \right) }{\partial \theta_S} \right|_{\theta_S = \theta_S } \right)^\top
  \end{aligned}
\end{equation}
Then, Equation~\ref{eqn:theta_t_second_term} becomes
\begin{equation}
  \begin{aligned}
    \underbrace{\frac{\partial \bar{\theta}_S' }{\partial \theta_T}}_{\ABS{S} \times \ABS{T}}
    &= - \eta_S \cdot \frac{\partial}{\partial \theta_T} \mathbb{E}_{\widehat{y}_{u} \sim T(x_u; \theta_T)} \Big[ \underbrace{g_S(\widehat{y}_{u})}_{\ABS{S} \times 1} \Big]
  \end{aligned}
\end{equation}
Since $g_S(\widehat{y}_{u})$ has no dependency on on $\theta_T$, except for via $\widehat{y}_{u}$, we can apply the REINFORCE equation~\cite{reinforce} to achieve
\begin{equation}
  \label{eqn:final_grad_theta_t}
  \begin{aligned}
    \underbrace{\frac{\partial \bar{\theta}_S^{(t+1)}}{\partial \theta_T}}_{\ABS{S} \times \ABS{T}}
    &= - \eta_S \cdot \frac{\partial}{\partial \theta_T} \mathbb{E}_{\widehat{y}_{u} \sim T(x_u; \theta_T)} \left[ g_S(\widehat{y}_{u}) \right] \\
    &= - \eta_S \cdot \mathbb{E}_{\widehat{y}_{u} \sim T(x_u; \theta_T)} \Big[
        \underbrace{g_S(\widehat{y}_{u})}_{\ABS{S} \times 1}
        \cdot
        \underbrace{\frac{\partial \log{P\left(\widehat{y}_{u} | x_{u}; \theta_T \right)}}{\partial \theta_T}}_{1 \times \ABS{T}}
    \Big] \\
    &= \eta_S \cdot \mathbb{E}_{\widehat{y}_{u} \sim T(x_u; \theta_T)} \Big[
        \underbrace{g_S(\widehat{y}_{u})}_{\ABS{S} \times 1}
        \cdot
        \underbrace{\frac{\partial \text{CE} \left( \widehat{y}_{u}, T(x_{u}; \theta_T) \right) }{\partial \theta_T}}_{1 \times \ABS{T}}
    \Big]
  \end{aligned}
\end{equation}
Here, the last equality in Equation~\ref{eqn:final_grad_theta_t} is is due to the definition of the cross-entropy loss, which is the negative of the log-prob term in the previous line.

Now, we can substitute Equation~\ref{eqn:final_grad_theta_t} into Equation~\ref{eqn:theta_t_chain_rule} to obtain
\begin{equation}
  \label{eqn:theta_t_assoc}
  \begin{aligned}
    \underbrace{\frac{\partial R}{\partial \theta_T}}_{1 \times \ABS{T}}
    &= \underbrace{\left. \frac{\partial \text{CE} \left(
      y_{l}, S\big (x_{l}; \bar{\theta}_S' \big) \right) }{\partial \theta_S} \right|_{\theta_S = \bar{\theta}_S' }}_{1 \times \ABS{S}}
    \cdot
    \underbrace{\frac{\partial \bar{\theta}_S'}{\partial \theta_T}}_{\ABS{S} \times \ABS{T}} \\
    &= \eta_S
       \cdot
       \underbrace{\left. \frac{\partial \text{CE} \left(
       y_{l}, S\big (x_{l}; \bar{\theta}_S' \big) \right) }{\partial \theta_S} \right|_{\theta_S = \bar{\theta}_S' }}_{1 \times \ABS{S}}
       \cdot
       ~\mathbb{E}_{\widehat{y}_{u} \sim T(x_u; \theta_T)} \Big[
        \underbrace{g_S(\widehat{y}_{u})}_{\ABS{S} \times 1}
        \cdot
        \underbrace{\frac{\partial \text{CE} \left( \widehat{y}_{u}, T(x_{u}; \theta_T) \right) }{\partial \theta_T}}_{1 \times \ABS{T}}
    \Big]
  \end{aligned}
\end{equation}
Finally, we use Monte Carlo approximation for every term in Equation~\ref{eqn:theta_t_assoc} using the sampled $\widehat{y}_{u}$. In particular, we approximate $\bar{\theta}_S'$ with the parameter obtained from $\theta_S$ by updating the student parameter on $(x_u, \widehat{y}_u)$, i.e., $\theta_S' = \theta_S - \eta_S \cdot \nabla_{\theta_S} \text{CE} \left( \widehat{y}_{u}, S(x_{u}; \theta_S) \right))$, and approximate the expected value in the second term with the same using $\widehat{y}_u$. With these approximation, we obtain the gradient $\nabla_{\theta_T} \mathcal{L}_u(\theta_T, \theta_S)$ from Equation~\ref{eqn:student}:
\begin{equation}
\label{eqn:theta_t_grad}
\begin{aligned}
  \nabla_{\theta_T} \mathcal{L}_l
  &= \eta_S
       \cdot
       \underbrace{\frac{\partial \text{CE} \left(
       y_{l}, S\big (x_{l}; \theta_S' \big) \right) }{\partial \theta_S}}_{1 \times \ABS{S}}
       \cdot
       \underbrace{\left( \left. \frac{\partial \text{CE} \left( \widehat{y}_{u}, S(x_{u}; \theta_S) \right) }{\partial \theta_S} \right|_{\theta_S = \theta_S } \right)^\top}_{\ABS{S} \times 1}
       \cdot
       \underbrace{\frac{\partial \text{CE} \left( \widehat{y}_{u}, T(x_{u}; \theta_T) \right) }{\partial \theta_T}}_{1 \times \ABS{T}} \\
  &= \underbrace{\eta_S \cdot \left(
  \Big( \nabla_{\theta_S'} \text{CE} \left( y_{l}, S(x_{l}; \theta_S' \right) \Big)^\top
  \cdot
  \nabla_{\theta_S} \text{CE} \left( \widehat{y}_{u}, S(x_{u}; \theta_S) \right)
  \right)}_{\text{A scalar $\coloneqq h$}}
  \cdot
  \nabla_{\theta_T} \text{CE} \left( \widehat{y}_{u}, T(x_{u}; \theta_T) \right)
\end{aligned}
\end{equation}

\section{\label{sec:mpl_uda}Pseudo Code for Meta Pseudo Labels with UDA}
In this section, we present the pseudo code for Meta Pseudo Labels where the teacher is trained with an extended objective to include the UDA loss. We emphasize that the UDA objective is applied \textit{on the teacher}, while the student still only learns from the pseudo labeled data given by the teacher. The pseudo code can be found in Algorithm~\ref{alg:training_procedure_uda}.
\begin{algorithm}[h]
\small
\SetAlgoLined
\DontPrintSemicolon
\SetKwInOut{Input}{Input}
\SetKwInOut{Output}{Output}
\SetCommentSty{itshape}
\SetKwComment{Comment}{$\triangleright$\ }{}
\textbf{Input:} Labeled data $x_{l}, y_{l}$ and unlabeled data $x_{u}$.

Initialize $\theta_T^{(0)}$ and $\theta_S^{(0)}$

\For{$t = 0$~\textbf{\emph{to}}~$N-1$}{
    Sample an unlabeled example $x_{u}$ and a labeled example $x_{l}, y_{l}$
    
    Sample a pseudo label $\widehat{y}_{u} \sim P(\cdot | x_{u}; \theta_T)$
    
    Update the student using the pseudo label $\widehat{y}_{u}$:$$\theta_{S}^{(t+1)} = \theta_{S}^{(t)} - \eta_S \left. \nabla_{\theta_S} \text{CE}(\widehat{y}_{u}, S(x_{u}; \theta_S) \right|_{\theta_S = \theta_{S}^{(t)}}$$
    
    Compute the teacher's feedback coefficient as in Equation~\ref{eqn:theta_t_grad}:$$h = \eta_S \cdot \left(
  \Big( \nabla_{\theta_S'} \text{CE} \left( y_{l}, S(x_{l}; \theta_S^{(t+1)} \right) \Big)^\top
  \cdot
  \nabla_{\theta_S} \text{CE} \left( \widehat{y}_{u}, S(x_{u}; \theta_S^{(t)}) \right)
  \right)$$
    
    Compute the teacher's gradient from the student's feedback:$$g^{(t)}_{T}
      =
        h
        \cdot
        \left.
          \nabla_{\theta_T} \text{CE}(\widehat{y}_{u}, T(x_{u}; \theta_T))
        \right|_{\theta_T = \theta_T^{(t)}}$$
    
    Compute the teacher's gradient on labeled data:$$g^{(t)}_{T, \text{supervised}} = \nabla_{\theta_T} \left. \text{CE}(y_{l}, T(x_{l}; \theta_T)) \right|_{\theta_T = \theta_T^{(t)}}$$
    
    Compute the teacher's gradient on the UDA loss with unlabeled data:$$g^{(t)}_{T, \text{UDA}} = \nabla_{\theta_T} \left. \text{CE}\Big(\text{StopGradient}(T(x_{l}); \theta_T), T(\text{RandAugment}(x_{l}); \theta_T) \Big) \right|_{\theta_T = \theta_T^{(t)}}$$
    
    Update the teacher:$$\theta_T^{(t+1)} = \theta_T^{(t)} - \eta_T \cdot \Big( g^{(t)}_{T} + g^{(t)}_{T, \text{supervised}} + g^{(t)}_{T, \text{UDA}}\Big) $$
    \label{alg:theta_t_update}
}
\Return{$\theta_S^{\text{(N)}}$} \hfill\Comment{Only the student model is returned for predictions and evaluations}
\captionof{algorithm}{\label{alg:training_procedure_uda}The Meta Pseudo Labels method, applied to a teacher trained with UDA~\cite{uda}.}
\end{algorithm}

\section{\label{sec:exp_details}Experimental Details}
In this section, we provide the training details for our experiments in Section~\ref{sec:exp} and Section~\ref{sec:large_scale_exp}.

\subsection{\label{sec:dataset_splits}Dataset Splits}
We describe how the datasets CIFAR-10-4K, SVHN-1K, and ImageNet-10\% in Section~\ref{sec:low_resource} are constructed. For CIFAR-10, we download the five training data batch files from CIFAR-10's official website.\footnote{CIFAR-10's official website: \url{www.cs.toronto.edu/~kriz/cifar.html}.} Then, we load all the images into a list of 50,000 images, keeping the order as downloaded. The fisrt 5,000 images are typically reserved for validation, so we remove them. The next 4,000 images are used as labeled data. For SVHN, we download the data from the \texttt{mat} files on SVHN's official site\footnote{SVHN's official website: \url{ufldl.stanford.edu/housenumbers/}.}, and follow the same procedure as with CIFAR-10. We note that this selection process leads to a slight imbalance in the class distribution for both CIFAR-10-4K and SVHN-1K, but the settings are the same for all of our experiments. For ImageNet, we follow the procedure in Inception's GitHub\footnote{Inception's GitHub, which also has the code to create ImageNet's training shards in \texttt{TFRecord}: \url{github.com/tensorflow/models/blob/master/research/inception/inception/data/download_and_preprocess_imagenet.sh}.}. This results in 1,024 training \texttt{TFRecord} shards of approximately the same size. The order of the images in these shards are deterministic. For ImageNet-10\%, we use the first 102 shards; for ImageNet-20\%, we use the first 204 shards; and so on. The last 20 shards, corresponding to roughly 25,000 images, are reserved for hyper-parameters tuning (used in Section~\ref{sec:resnet50} and Section~\ref{sec:large_scale_exp}).

\subsection{\label{sec:random_augment}Modifications of RandAugment~\cite{rand_augment}}
We modify a few data augmentation strategies as introduced by RandAugment~\cite{rand_augment}. Our modifications mostly target the SVHN dataset. In particular, we remove all rotations from the set of augmentation operations since rotation is a wrong invariance for digits such as 6 and 9. We also remove horizontal translations because they cause another wrong invariance for digits 3 and 8, e.g., when 8 is pushed half-outside the image and the remaining part looks like a 3. Table~\ref{tab:dataset_transformations} presents the transformations that we keep for our datasets.

\begin{table}[H]
  \small
  \centering
  \begin{tabular}{ll}
    \toprule
    \textbf{CIFAR-10} and \textbf{ImageNet} & \textbf{SVHN} \\
    \midrule
    AutoContrast & AutoContrast \\
    Brightness & Brightness \\
    Color & Color \\
    Contrast & Contrast \\
    Equalize & Equalize \\
    Invert & Invert \\
    Sharpness & Sharpness \\
    Posterize & Posterize \\
    Sample Pairing & Solarize \\
    Solarize & ShearX \\
    Rotate & ShearY \\
    ShearX & TranslateY \\
    ShearY & \\
    TranslateX & \\
    TranslateY & \\
    \bottomrule
  \end{tabular}
  \captionof{table}{\label{tab:dataset_transformations}Transformations that RandAugment uniformly samples for our datasets. We refer our readers to~\cite{auto_augment} for the detailed descriptions of these transformations.}
\end{table}

\subsection{\label{sec:additional_details}Additional Implementation Details}
To improve the stability of Meta Pseudo Labels, we use the following details in the Meta Pseudo Labels process.

\paragraph{Use cosine distance instead of dot product in Equation~\ref{eqn:theta_t_grad}.} The dot product $h$ in Equation~\ref{eqn:theta_t_grad} has a large value range, especially at the beginning of the Meta Pseudo Labels process. Thus, in order to stabilize training, we compute $h$ using the gradients' cosine distance.  This modification requires very little modification in our code.

We give two justifications why the use of cosine distance makes sense \textit{mathematically}. First, $h$ in Equation~\ref{eqn:theta_t_grad} is on a scalar which is multiplied with the teacher's gradient with respect to $\theta_T$. Changing dot product into cosine distance does not change the sign of $h$, and thus preserving the actions to increase or to decrease the probabilities of the sampled pseudo labels. Second, cosine distance's value range is much smaller than that of dot product, making the Meta Pseudo Labels updates more numerically stable. Specifically, the value range of cosine distance is $[-1, 1]$, while the value range of dot products, as observed in our experiments, is about $[-5 \times 10^4, 5 \times 10^4]$. This range also depends on the weight decay hyper-parameter.

Additionally, the dot product $h$, as shown in Equation~\ref{eqn:theta_t_grad} and as derived in Section~\ref{sec:theta_t_grad}, results from the application of the chain rule in a so-called bi-level optimization procedure. Bi-level optimization has been applied in some past work, such as Hyper Gradient Descent~\cite{hyper_gradient_descent}, which also replaces dot product with cosine distance to improve the numerical stability.

\paragraph{Use a baseline for $h$ in Equation~\ref{eqn:theta_t_grad}.} To further reduce the variance of $h$, we maintain a moving average $b$ of $h$ and subtract $b$ from $h$ every time we compute $g_T^{(t)}$ as in Equation~\ref{eqn:theta_t_grad}. This practice is also widely applied in Reinforcement Learning literature.

While using cosine distance is very crucial to maintain the numerical stability of Meta Pseudo Labels, using the moving average baseline only slightly improves Meta Pseudo Labels's performance. We suspect that not using the moving average baseline is also fine, especially when Meta Pseudo Labels can train for many steps without overfitting.

\subsection{\label{sec:hyper_parameters}Hyper-parameters}
\paragraph{Optimizers.} In all our experiments, the WideResNet-28-2 for CIFAR-10-4K and SVHN-1K and the ResNet-50 for ImageNet-10\% and full ImageNet are updated with Nesterov Momentum with default the momentum coefficient of 0.9. The networks' learning rate follow the cosine decay~\cite{cosine_lr}. Meanwhile, the EfficientNet-L2 and EfficientNet-B6-Wide for ImageNet+JFT are trained with RMSProp~\cite{rms_prop} and with an exponential decay learning rate. These are the default optimizers and learning rate schedules used for the architectures in their corresponding papers. We have only one substantial change of optimizer: when we finetune EfficientNet-L2 and EfficientNet-B6-Wide on the labeled data from ImageNet (see Section~\ref{sec:large_scale_exp}), we use the LARS optimizer~\cite{you2017large} with their default parameters, i.e., momentum 0.9 and learning rate 0.001, training for 20,000 steps with a batch size of 4,096. We finetune using this optimizer instead of SGD in Noisy Student~\cite{xie2020self} because unlike Noisy Student, the student model in Meta Pseudo Labels never trains directly on any labeled example, and hence can benefit from a more ``aggressive'' finetuning process with stronger optimiziers.

\paragraph{Numerical Hyper-parameters.}
To tune hyper-parameters, we follow~\cite{realistic_eval} and allow each method to have $128$ trials of hyper-parameters. When we tune, we let each model train for up to 50,000 steps. The optimal hyper-parameters are then used to run experiments that last for much more steps, as we report below. In our experiments with Meta Pseudo Labels, training for more steps typically leads to stronger results. We stop at 1 million steps for CIFAR-10-4K and SVHN-1K, and at 0.5 million steps for ImageNet because these are the standards from past papers.% Meanwhile, in our experiments with purely supervised learning, Pseudo-Labels, and UDA, training for more steps overfits the models, and we have to employ early stopping.

We report the hyper-parameters for our baselines and for Meta Pseudo Labels in Section~\ref{sec:exp} in Tables~\ref{tab:hparams_sup},~\ref{tab:hparams_uda},~\ref{tab:hparams_co_learning}. We note that our settings for UDA is different from originally reported by the original UDA paper~\cite{uda}. In their work, UDA~\cite{uda} use a much larger batch size for their UDA objective. In our implementation of UDA, we keep these batch sizes the same. This leads to a much easier implementation of data parallelism in our framework, TensorFlow~\cite{tensorflow} running on TPU big pods. To compensate for the difference, we train all UDA baselines for much longer than the UDA paper~\cite{uda}. During the training process, we also mask out the supervised examples with high confidence. Effectively, our UDA model receives roughly the same amount of training with labeled examples and unlabeled examples as the models in~\cite{uda}. We have also verified that on ImageNet-10\% with the augmentation policy from AutoAugment~\cite{auto_augment}, our UDA implementation achives $68.77\%$ top-1 accuracy, which is similar to $68.66\%$ that the UDA paper~\cite{uda} reported.
\begin{table}[H]
  \small
  \centering
  \begin{tabular}{llll}
    \toprule
    \textbf{Hyper-parameter} & \textbf{CIFAR-10} & \textbf{SVHN} & \textbf{ImageNet} \\
    \midrule
    Weight decay & 0.0005 & 0.001 & 0.0002 \\
    Label smoothing & 0 & 0 & 0.1 \\
    Batch normalization decay & 0.99 & 0.99 & 0.99 \\
    Learning rate & 0.4 & 0.05 & 1.28 \\
    Number of training steps & 50,000 & 50,000 & 40,000 \\
    Number of warm up steps & 2500 & 0 & 2000 \\
    Batch size & 1024 & 128 & 2048 \\
    Dropout rate & 0.4 & 0.5 & 0.2 \\
    \midrule
    Pseudo label threshold & 0.95 & 0.975 & 0.7 \\
    \bottomrule
  \end{tabular}
  \caption{\label{tab:hparams_sup}Hyper-parameters for supervised learning and Pseudo Labels.}
\end{table}

\begin{table}[H]
  \small
  \centering
  \begin{tabular}{llll}
    \toprule
    \textbf{Hyper-parameter} & \textbf{CIFAR-10} & \textbf{SVHN} & \textbf{ImageNet} \\
    \midrule
    Weight decay & 0.0005 & 0.0005 & 0.0002 \\
    Label smoothing & 0 & 0 & 0.1 \\
    Batch normalization decay & 0.99 & 0.99 & 0.99 \\
    Learning rate & 0.3 & 0.4 & 1.28 \\
    Number of training steps & 1,000,000 & 1,000,000 & 500,000 \\
    Number of warm up steps & 5,000 & 5,000 & 5,000 \\
    Batch size & 128 & 128 & 2048 \\
    Dropout rate & 0.5 & 0.6 & 0.25 \\
    UDA factor & 2.5 & 1 & 20 \\
    UDA temperature & 0.7 & 0.8 & 0.7 \\
    \bottomrule
  \end{tabular}
  \caption{\label{tab:hparams_uda}Hyper-parameters for UDA. Unlike originally done by the UDA paper~\cite{uda}, we do not use a larger batch size for the UDA objective. Instead, we use the same batch size for both the labeled objective and the unlabeled objective. This is to avoid instances where some particularly small batch sizes for the labeled objective cannot be split on our computational hardware.}
\end{table}

\begin{table}[H]
  \small
  \centering
  \begin{tabular}{lllll}
    \toprule
    & \textbf{Hyper-parameter} & \textbf{CIFAR-10} & \textbf{SVHN} & \textbf{ImageNet} \\
    \midrule
    \multirow{5}{*}{Common} & Weight decay & 0.0005 & 0.0005 & 0.0002 \\
    & Label smoothing & 0.1 & 0.1 & 0.1 \\
    & Batch normalization decay & 0.99 & 0.99 & 0.99 \\
    & Number of training steps & 1,000,000 & 1,000,000 & 500,000\\
    & Number of warm up steps & 2,000 & 2,000 & 1,000 \\
    \midrule
    \multirow{3}{*}{Student} & Learning rate & 0.3 & 0.15 & 0.8 \\
    & Batch size & 128 & 128 & 2048 \\
    & Dropout rate & 0.35 & 0.45 & 0.1 \\
    \midrule
    \multirow{5}{*}{Teacher} & Learning rate & 0.125 & 0.05 & 0.5 \\
    & Batch size & 128 & 128 & 2048 \\
    & Dropout rate & 0.5 & 0.65 & 0.1 \\
    & UDA factor & 1.0 & 2.5 & 16.0  \\
    & UDA temperature & 0.8 & 1.25 & 0.75 \\
    \bottomrule
  \end{tabular}
  \caption{\label{tab:hparams_co_learning}Hyper-parameters for Meta Pseudo Labels.}
\end{table}

\section{\label{sec:exp_analysis}More Detailed Analysis of Meta Pseudo Label's Behaviors}
We have seen in Section~\ref{sec:exp} and Section~\ref{sec:large_scale_exp} that Meta Pseudo Labels leads to strong performances on multiple image classification benchmarks. In this section, we provide further analysis of Meta Pseudo Labels and related baselines on more restricted and more controlled environments to provide better insights about Meta Pseudo Labels' behaviors.

\subsection{Visualizing the Contributions of Meta Pseudo Labels}
\begin{figure}[htb!]
  \centering
  \includegraphics[width=0.6\textwidth]{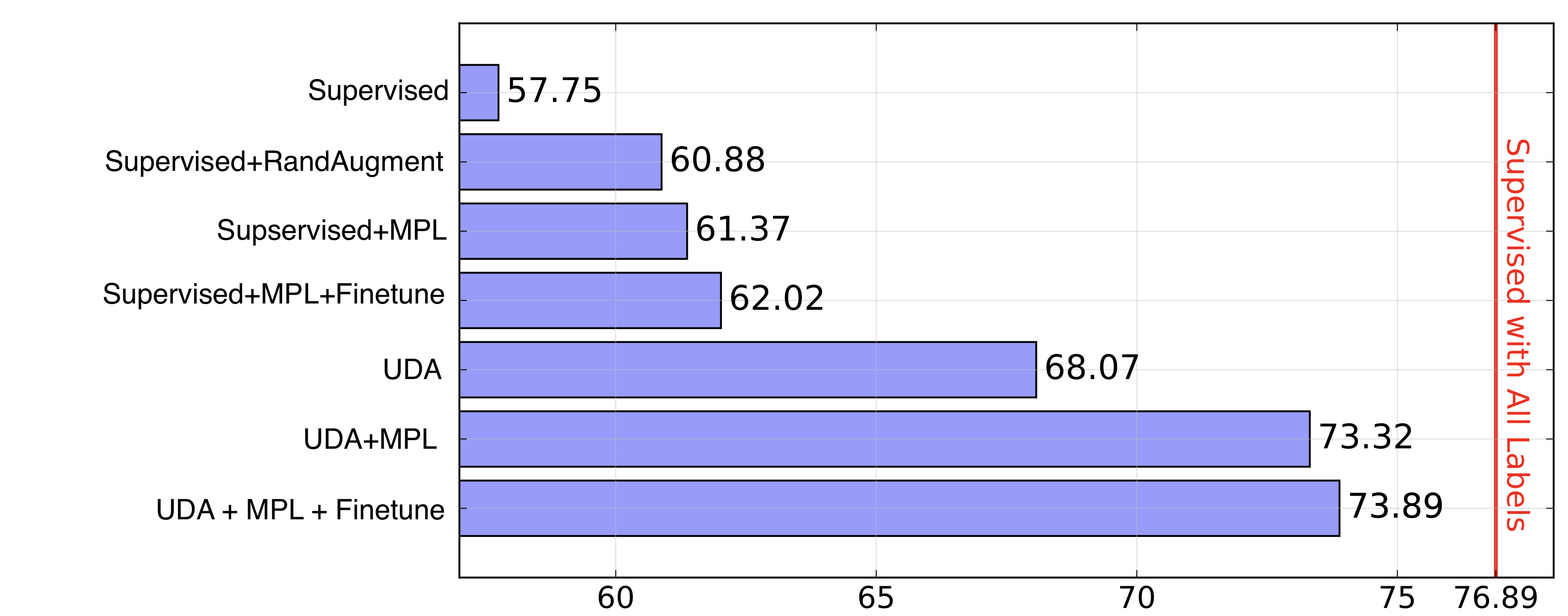}
  \captionof{figure}{\label{fig:ablation_imagenet}Breakdown of the gains of different components in Meta Pseudo Labels (MPL). The gain of Meta Pseudo Labels over UDA, albeit smaller than the gain of UDA over RandAugment, is significant as UDA is already very strong.}
\end{figure}

To understand the contributions of Meta Pseudo Labels (MPL), in Figure~\ref{fig:ablation_imagenet}, we visualize the relative gains of various methods on ImageNet-10\% (Section~\ref{sec:low_resource}). From the figure, we have two observations. First, for a purely supervised teacher, Meta Pseudo Labels outperforms RandAugment. We suspect this is because Meta Pseudo Labels is more effective form of regularization for the student. This is very crucial for ImageNet-10\%, where we only have about 128 images per class for each of the 1,000 classes. Second, UDA improves over Supervised+MPL+Finetune by 6.05\% in top-1 accuracy. This is in the same ballpark with the gain that UDA+MPL delivers above UDA, which is 5.25\%. As UDA's accuracy is already high, such improvement is very significant. Finally, finetuning only slightly improves over UDA+MPL. This extra performance boost is a unique advantage of Meta Pseudo Labels, since the student never directly learns from labeled data.

\subsection{\label{sec:mpl_reg}Meta Pseudo Labels Is An Effective Regularization Strategy}
The rest of this paper uses Meta Pseudo Labels as a semi-supervised learning method. In this section, we show that Meta Pseudo Labels can behave like an effective regularization method for supervised learning. This behavior can be achieved by making labeled data the same with unlabeled data in Figure~\ref{fig:mpl_process}. In this case, Meta Pseudo Labels can be seen as an adaptive form of Label Smoothing: the teacher generates soft labels on labeled data for the student, just like the way Label Smoothing smooths the hard labels to regularize the model. The main difference is that the policy in Label Smoothing is fixed, whereas the policy of the teacher in Meta Pseudo Labels is adaptive to enhance the student's performance.

To confirm the effect of Meta Pseudo Labels, we compare the method to Supervised Learning and Label Smoothing on CIFAR-10-4K and SVHN-1K. All models and settings are the same as in Section~\ref{sec:low_resource}, except that we do not use RandAugment and we restrict the unlabeled data to the same set of labeled data. We choose CIFAR-10-4K and SVHN-1K for this experiment because Label Smoothing is typically already used in ImageNet models. The results are shown in Table~\ref{tab:mpl_supervised}. As can be seen from the table, Meta Pseudo Labels achieves 83.71\% on CIFAR-10-4K and 91.89\% on SVHN-1K. Both of these are significantly better than the accuracy obtained by supervised learning with and without Label Smoothing. This shows the importance of feedback in Meta Pseudo Labels.
\begin{table}[h!]
\centering
\small
\resizebox{0.45\linewidth}{!}{ %
\begin{tabular}{lcc}
\toprule
&  \textbf{CIFAR-10-4K} & \textbf{SVHN-1K} \\
\midrule
Supervised & 82.14 $\pm$ 0.25 & 88.17 $\pm$ 0.47 \\
Label Smoothing & 82.21 $\pm$ 0.18 & 89.39 $\pm$ 0.25 \\
Meta Pseudo Labels & \textbf{83.71 $\pm$ 0.21} & \textbf{91.89 $\pm$ 0.14} \\
\bottomrule
\end{tabular}
} %
\caption{Meta Pseudo Labels can be used as a regularization method for supervised learning.}
\label{tab:mpl_supervised}
\end{table}

\subsection{\label{sec:confirmation_bias}Meta Pseudo Labels Is a Mechanism to Addresses the Confirmation Bias of Pseudo Labels}
\begin{figure}[H]
  \centering
  \includegraphics[width=0.4\textwidth]{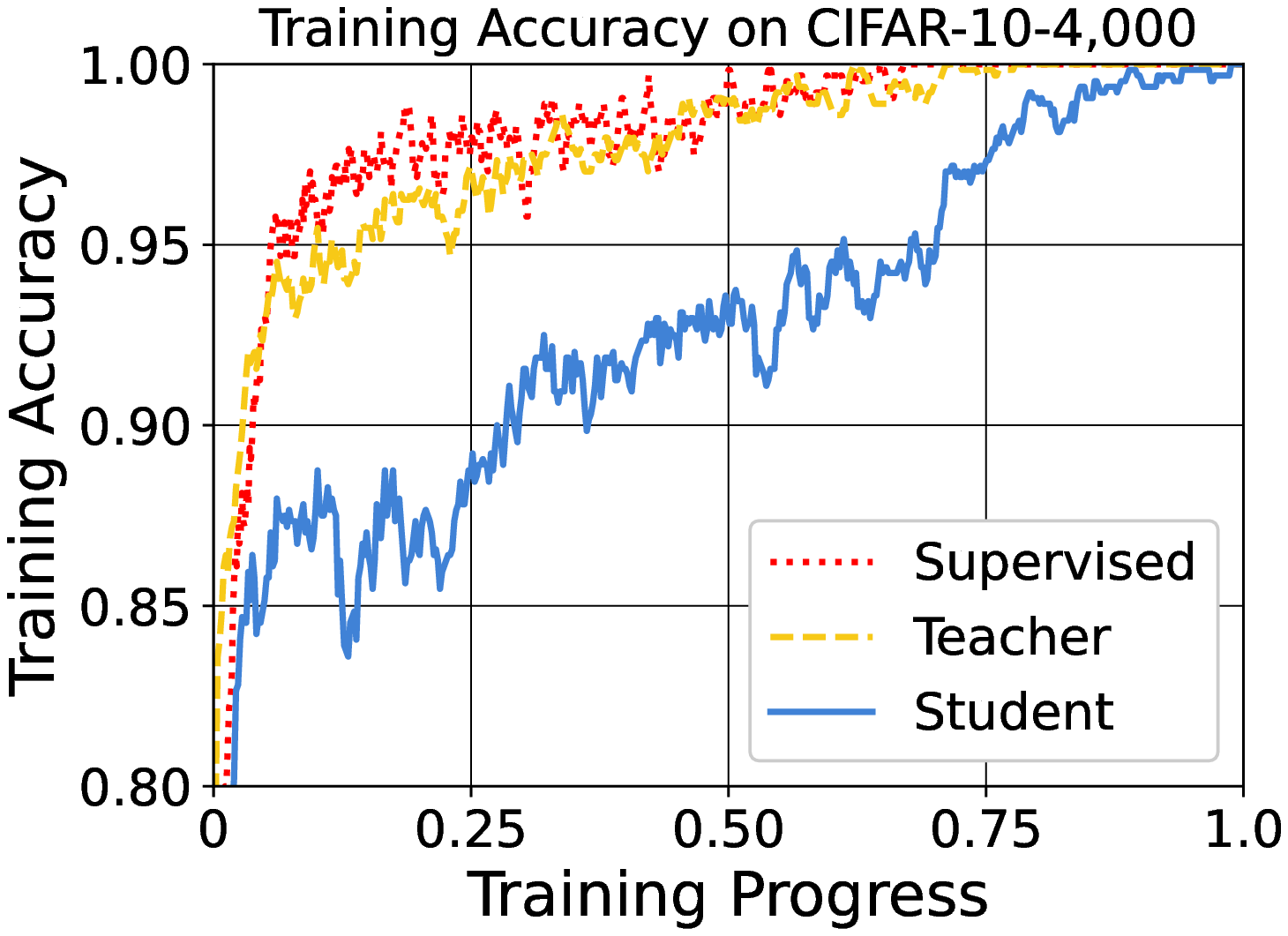}
  \includegraphics[width=0.4\textwidth]{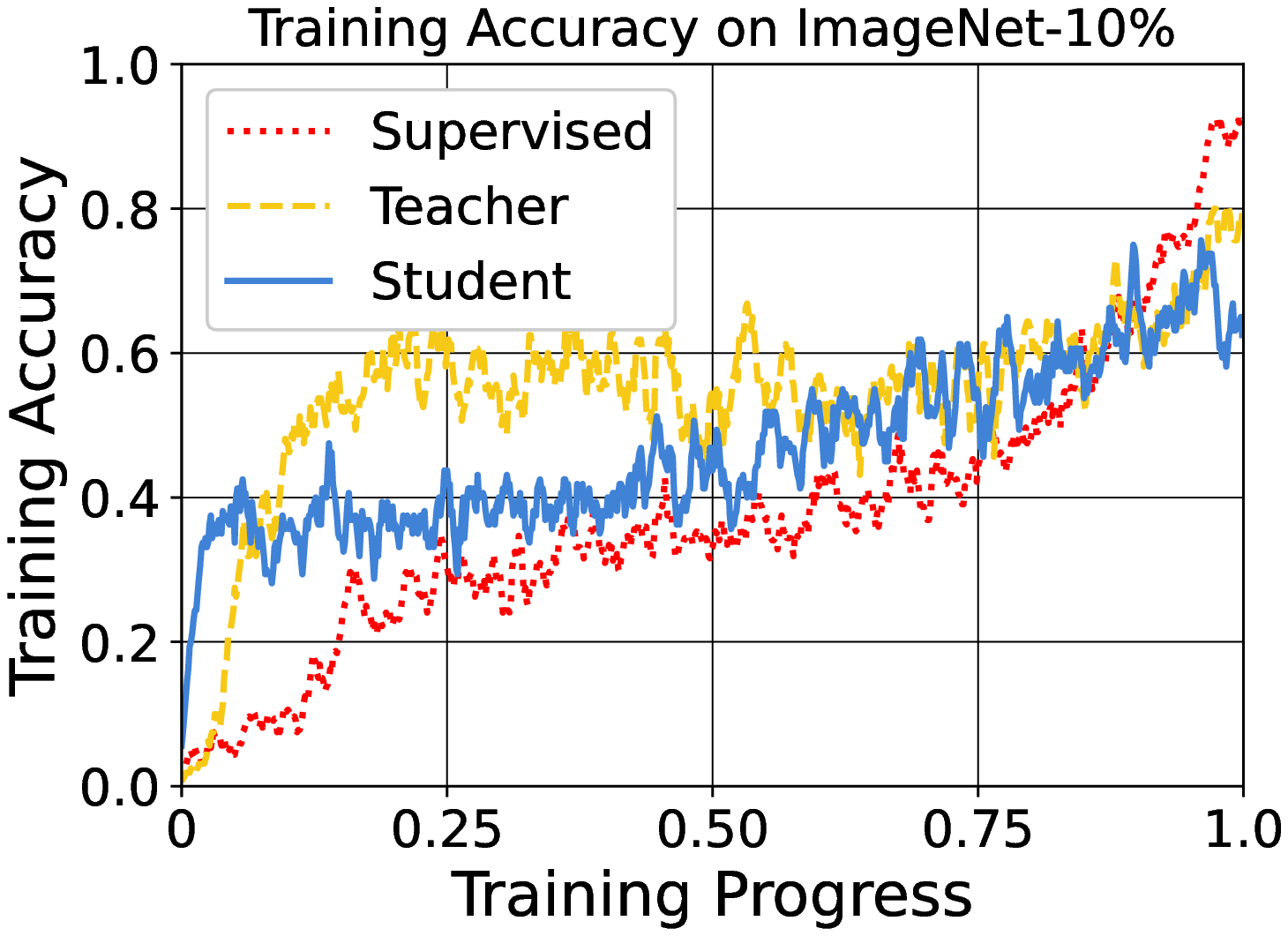}
  \captionof{figure}{\label{fig:train_acc}Training accuracy of Meta Pseudo Labels and of supervised learning on CIFAR-10-4,000 and ImageNet-10\%. Both the teacher and the student in Meta Pseudo Labels have lower training accuracy, effectively avoiding overfitting.}
\end{figure}

In this section, we show empirical evidence that Meta Pseudo Labels helps to address the teacher's confirmation bias~\cite{confirmation_bias} in Pseudo Labels. To this end, we analyze the \textit{training} accuracy of the teacher and the student in Meta Pseudo Labels from our experiments for CIFAR-10-4K and ImageNet-10\% in Section~\ref{sec:low_resource}. In Figure~\ref{fig:train_acc}, we plot the accuracy percentage at each training batch throughout the training process of a teacher and a student in Meta Pseudo Labels. We also plot the same data for a supervised model. From the figure, we have two observations:
\begin{itemize}
  \item On CIFAR-10-4K (Figure~\ref{fig:train_acc}-Left), the student's training accuracy in Meta Pseudo Labels is much lower that of the same network in Supervised Learning. As CIFAR-10-4K has very few labeled data, if the teacher converges quickly like in Supervised Learning, it will not generalize to the unlabeled data and hence will teach the student in inaccurate pseudo labels. In contrast, Figure~\ref{fig:train_acc}-Left shows that both the teacher and student in Meta Pseudo Labels converge much slower. To see this, note that in Meta Pseudo Labels, the student's \textit{training} accuracy is measured by how much it agrees with the teacher's pseudo labels. Therefore, the student in Meta Pseudo Labels having a lower training accuracy means that the student often disagrees with the pseudo labels that the teacher samples. This disagreement forces the teacher to constantly updates its weights to generate better pseudo labels, and makes it hard for the student to converge as the student has to learn from the teacher's changing pseudo labels. This behavior prevents both the teacher and the student from the premature convergence that causes the confirmation bias in Supervised Learning and Pseudo Labels.
  \item On ImageNet-10\% (Figure~\ref{fig:train_acc}-Right), the student also disagrees with the teacher's pseudo labels, as shown in the student's low training accuracy. Additionally, we observe that the teacher's training accuracy surges up faster than the supervised model's accuracy. We suspect that this is beneficial for the student learning, since ImageNet has 1,000 classes so in order to effectively teach the student to do well on the labeled dataset, the teacher has to become more accurate. Therefore, the feedback from the student is beneficial for the teacher's learn as well. This trend of high training accuracy only changes at the end of the training procedure, where the training accuracy of Supervised Learning surpasses those of the teacher and the student in Meta Pseudo Labels. From this last sign, we suspect that the supervised model has overfitted to the small set of labeled training examples in ImageNet-10\%, which will causes the confirmation bias if this supervised model is used to generate pseudo labels for another student model to learn from.
\end{itemize}

\subsection{Meta Pseudo Labels with Different Training Techniques for the Teacher}
In Sections~\ref{sec:exp} and Section~\ref{sec:large_scale_exp}, we have presented Meta Pseudo Labels results where the teacher is trained with UDA. In Table~\ref{tab:Meta Pseudo Labels_different_teachers}, we further show that on CIFAR-10-4K, Meta Pseudo Labels improves over different teachers trained with different techniques, including Pseudo Labels~\cite{pseudo_label}, Mixup~\cite{mixup}, and RandAugment. These results indicate that Meta Pseudo Labels is effective with all techniques. Additionally, the results suggest that better training techniques for the teacher tend to result in better students.
\begin{table}[H]
  \centering
  \resizebox{0.56\textwidth}{!}{
  \begin{tabular}{cccc}
    \toprule
    \textbf{Teacher} &
    \textbf{Pseudo-Labels} &
    \textbf{Mixup}~\cite{mixup} &
    \textbf{RandAugment} \\
    \midrule
    -Meta Pseudo Labels &
    $83.79 \pm 0.11$ &
    $84.20 \pm 0.15$ &
    $85.53 \pm 0.25$ \\
    +Meta Pseudo Labels &
    \textbf{84.11 $\pm$ 0.07} &
    \textbf{84.81 $\pm$ 0.19} &
    \textbf{87.55 $\pm$ 0.14} \\
    \bottomrule
  \end{tabular}
  }
  \caption{\label{tab:Meta Pseudo Labels_different_teachers}Meta Pseudo Labels's accuracy for WideResNet-28-2 on CIFAR-10-4,000, where the teacher is trained with different techniques. All numbers are $\text{mean} \pm \text{std}$ over 10 runs.}
\end{table}

\subsection{Meta Pseudo Labels with Different Amounts of Labeled Data}
\begin{figure}[H]  % imagenet_224_v_shards
  \centering
  \includegraphics[width=0.5\textwidth]{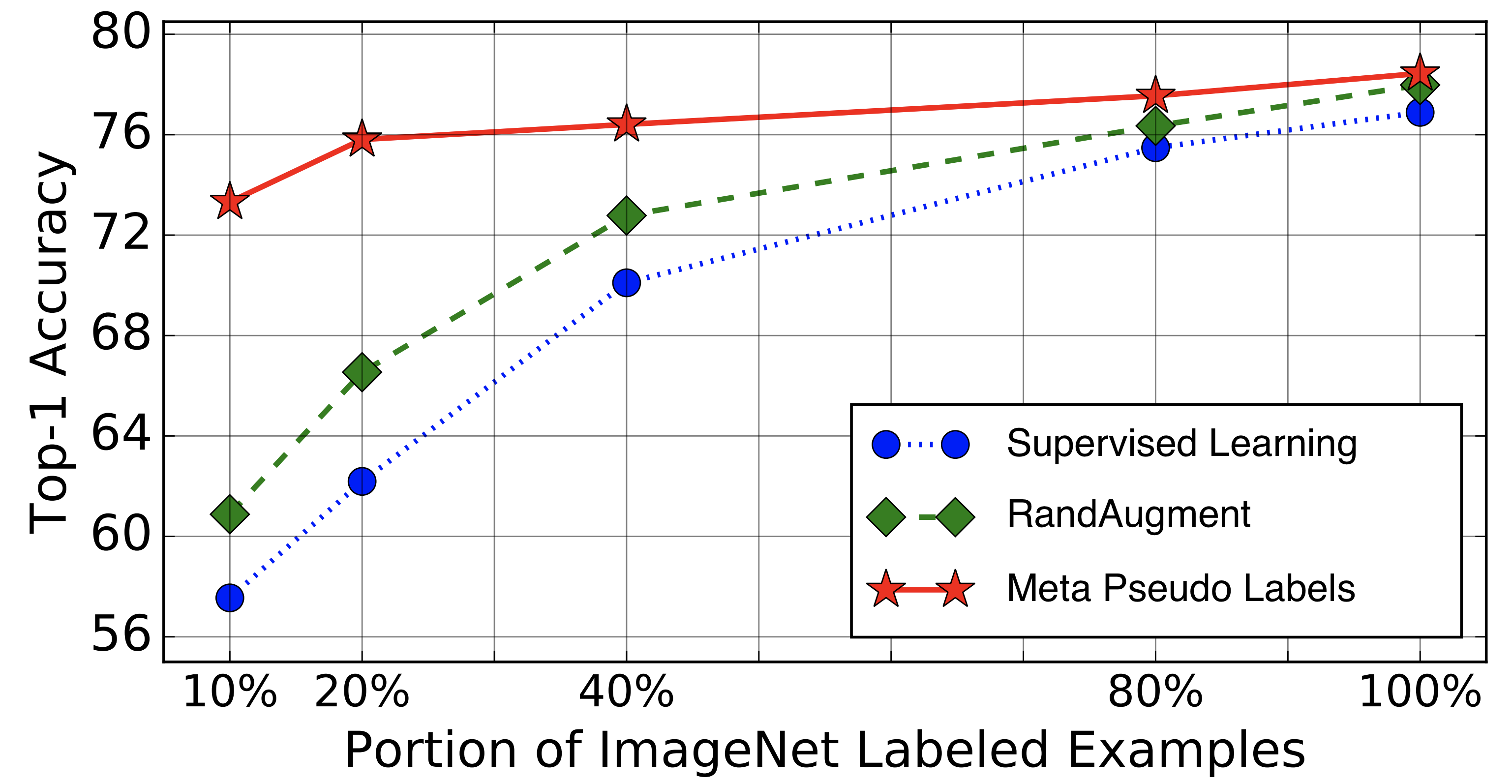}
  \captionof{figure}{\label{fig:multi_labels_imagenet}Performance of Supervised Learning, RandAugment, and Meta Pseudo Labels at different amounts of labeled examples.}
\end{figure}

We study how much Meta Pseudo Labels improves as more labeled data becomes available. To this end, we experiment with 10\%, 20\%, 40\%, 80\%, and 100\% of the labeled examples in ImageNet. We compare Meta Pseudo Labels with supervised learning and RandAugment. We plot the results in Figure~\ref{fig:multi_labels_imagenet}. From the figure, it can be seen that Meta Pseudo Labels delivers substantial gains with less data, but plateaus as more labeled data becomes available. This result suggests that Meta Pseudo Labels is more effective for low-resource image classification problems.

\section{\label{sec:reduced_mpl}Results with An Economical Version of Meta Pseudo Labels}
Meta Pseudo Labels requires storing both the teacher model and the student model in memory. For model architectures with a large memory footprint, such as EfficientNet-L2 and EfficientNet-B6-Wide in our experiments, this memory footprint exceeds 16G of available memory in our accelerators. While we have implemented a hybrid data-model parallelism in Section~\ref{sec:large_scale_exp} which allows us to run Meta Pseudo Labels with large model architectures, the tradeoff is a slow and expensive training procedure. To allow a more efficient training of large models with Meta Pseudo Labels, we design a more economical alternative to instantiate the teacher, termed Reduced Meta Pseudo Labels.

In Reduced Meta Pseudo Labels, we first train a large teacher model $T$ to convergence. Next, we use $T$ to pre-compute all target distributions for the student's training data. Importantly, until this step, the student model has not been loaded into memory, effectively avoiding the large memory footprint of Meta Pseudo Labels. Then, we parameterize a reduced teacher $T'$ as a small and efficient network, such as a multi-layered perceptron (MLP), to be trained the along with student. This reduced teacher $T'$ takes as input the distribution predicted by the large teacher $T$ and outputs a calibrated distribution for the student to learn. Intuitively, Reduced Meta Pseudo Labels works reasonably well because the large teacher $T$ is reasonably accurate, and hence many actions of the reduced teacher $T'$ would be close to an identity map, which can be handled by an MLP. Meanwhile, Reduced Meta Pseudo Labels retains the benefit of Meta Pseudo Labels, as the teacher $T'$ can still adapt to the learning state of the student $\theta_T$.

To evaluate whether Meta Pseudo Labels~can scale to problems with a large number of labeled examples, we now turn to full labeled sets of CIFAR-10, SVHN and ImageNet. We use out-of-domain unlabeled data for CIFAR-10 and ImageNet. We experiment with Reduced Meta Pseudo Labels~whose memory footprint allows our large-scale experiments. We show that the benefit of Meta Pseudo Labels, \ie~having a teacher that adapts to the student's learning state throughout the student's learning, stil extends to large datasets with more advanced architectures and out-of-domain unlabeled data. 

\paragraph{Model Architectures.} For our student model, we use EfficinetNet-B0 for CIFAR-10 and SVHN, and use EfficientNet-B7 for ImageNet. Meanwhile, our teacher model is a small 5-layer perceptron, with ReLU activation, and with a hidden size of 128 units for CIFAR-10 and of 512 units for ImageNet.

\paragraph{Labeled Data.} Per standard practices, we reserve 4,000 examples of CIFAR-10, 7,300 examples from SVHN, and 40 data shards of ImageNet for hyper-parameter tuning. This leaves about 45,000 labeled examples for CIFAR-10, 65,000 labeled examples for SVHN, and 1.23 million labeled examples for ImageNet. As in Section~\ref{sec:low_resource}, these labeled data serve as both the validation data for the student and the pre-training data for the teacher.

\paragraph{Unlabeled Data.} For CIFAR-10, our unlabeled data comes from the TinyImages dataset which has 80 million images~\cite{tinyimages}. For SVHN, we use the extra images that come with the standard training set of SVHN which has about 530,000 images. For ImageNet, our unlabeled data comes from the YFCC-100M dataset which has 100 million images~\cite{yfcc100m}. To collect unlabeled data relevant to the tasks at hand, we use the pre-trained teacher to assign class distributions to images in TinyImages and YFCC-100M, and then keep $K$ images with highest probabilities for each class. The values of $K$ are 50,000 for CIFAR-10, 35,000 for SVHN, and 12,800 for ImageNet.

\paragraph{Baselines.} We compare Reduced Meta Pseudo Labels~to NoisyStudent~\cite{xie2020self}, because it can be directly compared to Reduced Meta Pseudo Labels. In fact, the \textit{only} difference between NoisyStudent and Reduced Meta Pseudo Labels~is that Reduced Meta Pseudo Labels~has a teacher that adapts to the student's learning state.

\begin{table}[htb!]
\centering
\resizebox{0.57\textwidth}{!}{%
  \begin{tabular}{lccc}
    \toprule
    \textbf{Methods} & \textbf{CIFAR-10} & \textbf{SVHN} & \textbf{ImageNet} \\
    \midrule
    Supervised & $97.18 \pm 0.08$ & $98.17 \pm 0.03$ & $84.49 / 97.18$ \\
    NoisyStudent & $98.22 \pm 0.05$ & \textbf{98.71 $\pm$ 0.11} & $85.81 / 97.53$ \\
    \midrule
    Reduced Meta Pseudo Labels & \textbf{98.56 $\pm$ 0.07} & \textbf{98.78 $\pm$ 0.07} & \textbf{86.87$/$98.11} \\
    \bottomrule
  \end{tabular}
}%
\captionof{table}{\label{tab:reducedmpl}Image classification accuracy of EfficientNet-B0 on CIFAR-10 and SVHN, and EfficientNet-B7 on ImageNet. Higher is better. CIFAR-10 results are mean $\pm$ std over 5 runs, and ImageNet results are top-1/top-5 accuracy of a single run. All numbers are produced in our codebase and are controlled experiments.}
\end{table}

\paragraph{Results.} As presented in Table~\ref{tab:reducedmpl}, Reduced Meta Pseudo Labels~outperforms NoisyStudent on both CIFAR-10 and ImageNet, and is on-par with NoisyStudent on SVHN. 
In particular, on ImageNet, Meta Pseudo Labels~ with EfficientNet-B7 achieves a top-1 accuracy of 86.87\%, which is 1.06\% better than the strong baseline NoisyStudent. On CIFAR-10, Meta Pseudo Labels~ leads to an improvement of 0.34\% in accuracy on NoisyStudent, marking a 19\% error reduction. 

% We suspect there are two reasons to explain why the gain of Reduced Meta Pseudo Labels~on CIFAR-10 and SVHN are not significant. 
For SVHN, we suspect there are two reasons of why the gain of Reduced Meta Pseudo Labels~is not significant. 
First, NoisyStudent already achieves a very high accuracy. Second, the unlabeled images are high-quality, which we know by manual inspection. Meanwhile, for many ImageNet categories, there are not sufficient images from YFCC100M, so we end up with low-quality or out-of-domain images. On such noisy data, Reduced Meta Pseudo Labels's adaptive adjustment becomes more crucial for the student's performance, leading to more significant gain.
\end{document}